\documentclass[final]{cvpr}
\def\cvprPaperID{XXXX} %
\def\confYear{}
\usepackage{times}
\usepackage{epsfig}
\usepackage{graphicx}
\usepackage{amsmath}
\usepackage{amssymb}

\usepackage{comment}
\usepackage{enumitem}
\usepackage{color}
\usepackage[nocompress,sort]{cite}
\usepackage{multicol}
\usepackage{widetable}

\newcommand{\Fig}{Fig.\@\xspace}

\newcommand{\Sec}{Sec.\@\xspace}
\newcommand{\Tab}{Tab.\@\xspace}
\newcommand{\Eq}{Eq.\@\xspace}

\newcommand\customsubsection[1]{\vspace{0.4em}\noindent\textbf{#1}}

\usepackage{array}
\usepackage{multirow}
\usepackage{booktabs}
\usepackage{colortbl}
\usepackage{xcolor}
\usepackage{etoolbox,siunitx}
\robustify\bfseries
\definecolor{table_gray}{rgb}{0.85, 0.85, 0.85}

\usepackage{float} %
\usepackage{dsfont}
\usepackage{enumitem}
\usepackage{microtype}
\setlist[enumerate]{itemsep=0mm}
\newcommand{\+}[1]{\ensuremath{\mathbf{#1}}}
\DeclareMathOperator{\argmin}{argmin}

\newcommand\innerprod[2]{\langle #1\ , #2 \rangle}

\usepackage{pifont}
\newcommand{\cmark}{\ding{51}}
\newcommand{\xmark}{\ding{55}}

\usepackage{url}

\usepackage[strict]{changepage}

\usepackage[pagebackref=true, pdfa]{hyperref}

\definecolor{linkcolour}{rgb}{0.949,0.419,0.541}    %
\definecolor{citecolour}{rgb}{0.000,0.600,0.200}  %
\definecolor{urlcolour} {rgb}{0.078,0.337,0.564}    %

\hypersetup{
    bookmarksopen=false,
    bookmarksnumbered=true,
    pdfpagemode=UseOutlines,  %
    colorlinks=true,          %
    linkcolor=linkcolour,     %
    anchorcolor=black,        %
    citecolor=citecolour,     %
    filecolor=urlcolour,      %
    menucolor=blue,           %
    urlcolor=urlcolour,       %
    breaklinks=true,          %
    pdfstartview={FitBH},
    pdfview={FitBH},
}

\usepackage[labelfont=bf,
            font=small,
            skip=0.5ex,
            format=plain]{caption}

\begin{document}

\title{\vspace*{-2ex} DeepVideoMVS: Multi-View Stereo on Video\\with Recurrent Spatio-Temporal Fusion \vspace*{-1ex}}

\author{\hspace{-1.4em}
        Arda Düzçeker$^1$,
        Silvano Galliani$^2$,
        Christoph Vogel$^2$,
        Pablo Speciale$^2$,
        Mihai Dusmanu$^1$,
        Marc Pollefeys$^{1,2}$\\
        \hspace{-1.5em}$^1$Department of Computer Science, ETH Zurich\\
        \hspace{-1.5em}$^2$Microsoft Mixed Reality \& AI Zurich Lab}

\twocolumn[{%
\renewcommand\twocolumn[1][]{#1}%
\maketitle
\begin{center}
\vspace{-2ex}
\includegraphics[width=\linewidth]{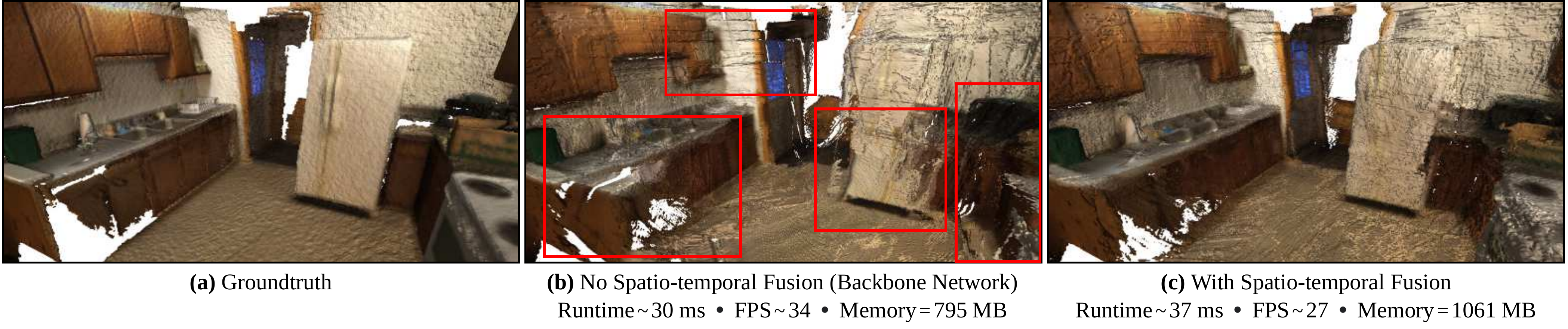}
\vspace{-2.5ex}
\captionof{figure}
    {
        \textls[-22] 
        {
        3D reconstructions of a scene from ScanNet~\cite{dai_scannet_2017}.
        Extending our stereo backbone with our proposed spatio-temporal fusion module improves the temporal consistency and accuracy of the predicted depth maps,
        leading to better reconstructions with negligible computational overhead. 
        Runtime is per forward pass on an NVIDIA GTX 1080Ti with image size $320\times256$.
        }
    }
\label{figure_teaser}
\vspace{2ex}
\end{center}
}]
\begin{abstract}
\vspace*{-0.5ex}
We propose an online multi-view depth prediction approach on posed video streams, where the scene geometry information computed in the previous time steps is propagated to the current time step in an efficient and geometrically plausible way. The backbone of our approach is a real-time capable, lightweight encoder-decoder that relies on cost volumes computed from pairs of images. We extend it by placing a ConvLSTM cell at the bottleneck layer, which compresses an arbitrary amount of past information in its states. The novelty lies in propagating the hidden state of the cell by accounting for the viewpoint changes between time steps. At a given time step, we warp the previous hidden state into the current camera plane using the previous depth prediction.
Our extension brings only a small overhead of computation time and memory consumption, while improving the depth predictions significantly. As a result, we outperform the existing state-of-the-art multi-view stereo methods on most of the evaluated metrics in hundreds of indoor scenes while maintaining a real-time performance. Code available: \small{\url{https://github.com/ardaduz/deep-video-mvs}}
\end{abstract}

\vspace*{-2ex}
\section{Introduction}

\noindent Obtaining dense 3D information about the environment is key for a wide range of applications such as navigation for autonomous vehicles (\eg, robots, drones~\cite{Scaramuzza2020}), mixed reality~\cite{Microsoft2019Announcing, Facebook2019Inside, Google2019Google, MagicLeap2019What}, 3D modelling and industrial control.
Compared to active depth sensing with LiDAR~\cite{Thinal2020}, %
time-of-flight~\cite{10.5555/2414778} or structured-light cameras~\cite{Fofi2004_structured_light}, 
camera-based passive sensing has the advantage of %
being energy and cost efficient, compact in size and operating in a wide range of conditions~\cite{ma_sparse--dense_2018}.
Among passive depth sensing approaches, monocular systems can offer highly mobile, low-maintenance solutions, while stereo devices often require baseline sizes that are infeasible for mobile devices~\cite{sinha_deltas_2020}.

One common denominator of the aforementioned applications is that the data is acquired as a video stream 
instead of sparse instances in time, and the depth is often reconstructed for selected keyframes. %
In this work, we assume a calibrated camera and known poses between acquisitions %
and focus on the dense depth recovery for each keyframe.
Such pose information can, for instance, be obtained through visual-inertial odometry techniques~\cite{bloesch2017iterated,Qin2017Vins},
which are readily available in mobile platforms (\eg~Apple ARKit and Android ARCore)~\cite{hou_multi-view_2019} or mixed reality headsets such as Microsoft HoloLens.
The presence of camera poses enables the computation of triangulation-based metric reconstructions, 
as opposed to the popular learning-based single image depth prediction 
methods~\cite{eigen_depth_2014, fu_deep_2018, alhashim_high_2019, yin_enforcing_2019, lee_big_2020} 
that have been extended to video~\cite{patil_dont_2020, leal-taixe_temporally_2019, zhang_exploiting_2019, wang_recurrent_2019}. 
Finally, the real-time aspect of the applications and the potential of an on-device solution,  
imply targeting a lightweight online multi-view stereo (MVS) system that is memory and compute efficient.
Therefore, similar to~\cite{hou_multi-view_2019, liu_neural_2019}, 
we specifically aim to harness the advantages of video, 
with limited variation in viewpoint at consecutive time steps, 
instead of pursuing unstructured MVS.

In this work, we present a framework that can extend many existing MVS methods, such that, when processing video streams, 
partial scene geometry information from the past contributes to %
the prediction at the current time step, 
leading to improved and consistent depth outputs. %
We use convolutional long short-term memory (ConvLSTM)~\cite{shi_convolutional_2015} and a hidden state 
propagation scheme to achieve such information flow in the latent space. %
Our approach is influenced by~\cite{hou_multi-view_2019}, where latent cost volume encodings are weakly coupled through a Gaussian Process, 
and by \cite{patil_dont_2020} where latent encodings of image features and sparse depth cues are fused through 
ConvLSTM to achieve temporally consistent depth predictions. 
However, we leverage geometry and explicitly account for the implications of perspective projection when propagating the latent encodings. %

Our key contributions are as follows:
\textbf{(\romannumeral 1)}
We propose a compact, cost-volume-based, stereo depth estimation network that solely relies on 2D convolutions to obtain real-time and memory efficient processing.
\textbf{(\romannumeral 2)}
We extend our model with a ConvLSTM cell, an explicit hidden state propagation scheme, 
and a training/inference strategy to enable spatio-temporal information flow.
This extension significantly improves the depth estimation accuracy, while creating only a small computational overhead, \cf~\Fig\ref{figure_teaser}.
\textbf{(\romannumeral 3)}
We set a new state-of-the-art on ScanNet~\cite{dai_scannet_2017}, 7-Scenes~\cite{glocker_real-time_2013}, TUM RGB-D~\cite{sturm_benchmark_2012} and ICL-NUIM~\cite{handa_benchmark_2014}, \cf~Tab.~\ref{table_depth_prediction_quantitative}, while obtaining the lowest runtime and a small memory footprint, \cf~\Fig~\ref{figure_inference_memory_vs_accuracy}.

\section{Related Work}
\noindent The most common representation for learning-based MVS is depth map, where 3D reconstruction is performed at a later stage, if needed. 
Compared to direct inference in 3D space with either an explicit voxel discretization~\cite{kar_learning_2017, ji_surfacenet_2017},
point-based representations~\cite{chen_point-based_2019}, implicit neural scene representations~\cite{mildenhall2020nerf, Mescheder_2019_CVPR}, or direct regression of a truncated signed distance function (TSDF)~\cite{murez_atlas_2020}, 
depth map representations are more versatile and can be used for various other tasks in addition to 3D reconstruction. 
Furthermore, this simple 2D representation appears to be more memory efficient and 
capable of delivering real-time information. %

\customsubsection{Depth Map Prediction with Learned MVS}. 
Most learning-based MVS methods follow traditional plane-sweeping~\cite{collins_space-sweep_1996, gallup_real-time_2007} 
to generate a cost volume from a designated reference and a measurement frame. %
On the one hand, methods such as MVSNet~\cite{yao_mvsnet_2018} and DPSNet~\cite{im_dpsnet_2019} build 4D feature volumes~\cite{kendall_end--end_2017}, and regularize the feature volumes by employing 3D convolutions, a process that delivers high accuracy, but is computationally demanding. 
On the other hand, MVDepthNet~\cite{wang_mvdepthnet_2018} and~\cite{yee_fast_2019}, directly generate 3D volumes by computing traditional cost measures on image features or RGB values. 
This allows basing the network architecture on 2D convolutions, 
which are faster than the 3D counterpart, and better suited for real-time applications.
As a compromise, predetermined cost measures decimate the color and feature information, which is often tackled by providing the reference image to the network, in addition to the cost volume.
Targeting real-time performance, we propose a similar, especially lightweight network based on these principles as our backbone.
A notable exception to cost-volume-based MVS is DELTAS~\cite{sinha_deltas_2020}, which learns to detect and triangulate interest points in the input images, 
and densifies the sparse set of 3D points to produce dense depth maps. 
However, the different approach of DELTAS can still be extended by our fusion framework.

\customsubsection{Depth Estimation from Video}. 
In typical video data, successive frames are strongly correlated and spatially close in the 3D space. 
Several single image depth prediction methods~\cite{patil_dont_2020, leal-taixe_temporally_2019, zhang_exploiting_2019} and stereo depth estimation methods~\cite{Zhong_2018_ECCV, Zhan_2018_CVPR} have shown that modelling the temporal relations or introducing optimization constraints among frames can improve the depth/disparity predictions.

In \cite{leal-taixe_temporally_2019}, only monocular image sequences are input and the temporal relations 
between image representations are modeled by placing many ConvLSTM cells in a single image depth estimation network.
Similarly,~\cite{zhang_exploiting_2019} employs ConvLSTM to model the relations between the successive frames, 
and extend their model with a generative adversarial network~\cite{goodfellow_generative_2014} and a temporal loss, 
to enforce consistency among video frames. 
Despite achieving temporal consistency and visually pleasant results,~\cite{leal-taixe_temporally_2019} and~\cite{zhang_exploiting_2019}
cannot ensure geometric correctness due to the lack of geometric foundation based on image measurements.
In~\cite{patil_dont_2020}, input frames and sparse depth cues are encoded together,
then relations between consecutive latent encodings are modeled through a ConvLSTM. Finally, the dense depth predictions are output by a decoder.\cite{Zhong_2018_ECCV} propose an unsupervised learning setting on stereo videos. They take the stereo image sequences as input and exploit the temporal dynamics by placing two separate ConvLSTMs in their network to predict more accurate disparity maps. ~\cite{Zhan_2018_CVPR} establishes temporal and stereo constraints on consecutive frames of the stereo sequence to improve the joint pose and depth estimations in their unsupervised framework.
In contrast, known camera poses, \eg MVS, do not only enable triangulated metric measurements from arbitrary set of frames, but also the transfer
of the past scene geometry encodings into the current view geometry, which improves the accuracy in our model.

Lately, video input has been leveraged also in learning-based MVS. 
GP-MVS~\cite{hou_multi-view_2019} extends MVDepthNet by introducing a Gaussian Process (GP)
at the bottleneck between the cost volume encoder-decoder of MVDepthNet.
The method constructs a GP prior kernel from a similarity measure between the known camera poses. 
This introduces soft constraints at the bottleneck layer and encourages the model to produce similar latent encodings for the frames that have similar poses. They also propose an inference scheme that evolves the GP in state-space for online operation.
In comparison, we fuse the latent encodings through learned convolution filters, after perspective correction, whereas GP-MVS performs non-parametric fusion constrained by pose priors, without any perspective correction.
Neural RGBD~\cite{liu_neural_2019} estimates a depth probability distribution per pixel, 
and propagates the resulting probability volume through time. 
As successive video frames are processed, the depth probability volumes are aggregated under a Bayesian filtering framework, 
which contributes to the confidence of a pixel's depth hypothesis.
In contrast, we propose to do the fusion at an earlier stage and operate on the 
latent encodings at the bottleneck, instead of explicitly working on probability volumes.

\section{Method}
\begin{figure*}[t]
\centering
{\includegraphics[clip, trim={0 0 0 1.05cm}, width=0.85\linewidth]{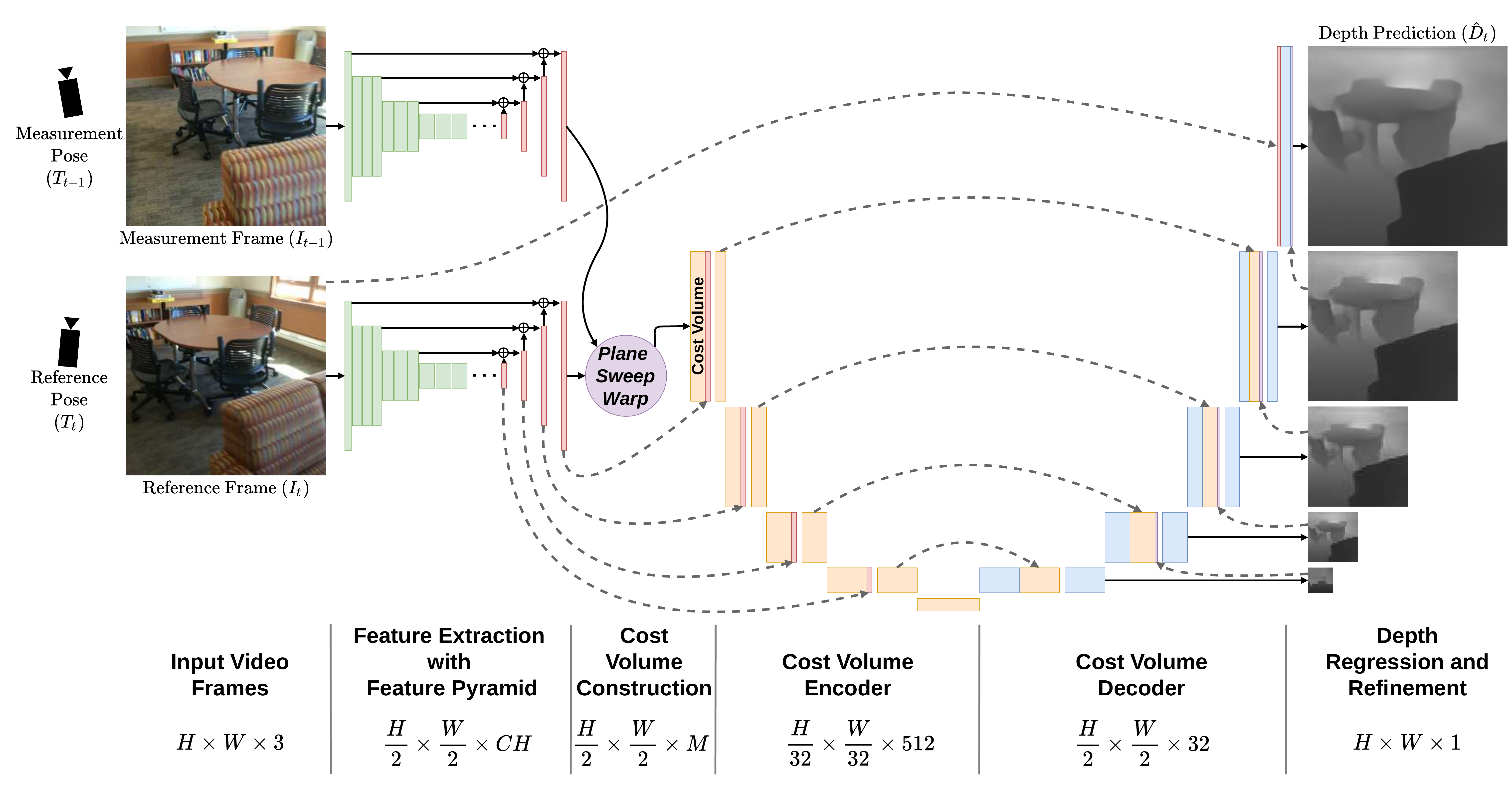}}
\caption{
    Sketch of our pair network that consists of five main parts. 
    First is the shared feature extraction with feature pyramid network to acquire feature maps from input images. 
    Second is the cost volume construction module that builds a 3D cost volume using the extracted features. 
    Then, an encoder-decoder network regularizes the cost volume. 
    We allow the encoder to take cues from extracted feature maps by placing skip connections from the feature extraction module. 
    Finally, the model regresses depth maps at multiple resolutions including the input image resolution after a small refinement block.
    }
\label{figure_pair_network}
\end{figure*}
\noindent For an online multi-view stereo system (with no temporal delay) that works on posed-video input, the supervised learning problem can be formulated as \vspace{-0.5ex}
\begin{equation}
\widehat{\+D}_{t} \!=\! f_{\theta}(\+I_t,\+I_{t-1},\ldots,\+I_{t-\delta},\+T_t,\+T_{t-1},\ldots,\+T_{t-\delta},\+K)\vspace{-0.5ex},
\label{equation_mvs_black_box}
\end{equation}
\begin{equation}
\theta^{\star} = \argmin_{\theta} \ l\left(\widehat{\+D}_{t},\ \+D_{t}\right).\vspace{-0.5ex}
\end{equation}
The task is to learn a predictor $f_{\theta}$ with a set of learnable parameters $\theta$ that can infer a depth map $\widehat{\+D}_{t}$, 
which is as close as possible to the groundtruth depth $\+D_t$~\cite{laga_survey_2019} by having access to the camera intrinsic matrix $\mathbf{K}$, 
images $\+I$ and camera poses $\+T\in{SE(3)}$, corresponding to the current time step $t$ and a number of $\delta$ previous time steps 
$t-1, \ldots, t-\delta$. \linebreak
$l(\cdot, \cdot)$ is the loss between the inferred and the groundtruth depth map, which we want to minimize over a large dataset. 

Our approach is discussed in two sections. 
\Sec~\ref{subsection_pair_network} introduces a cost-volume-based, stereo depth prediction network, \cf~\Fig~\ref{figure_pair_network}. 
This lightweight model %
serves as the backbone structure that we build upon. 
\Sec~\ref{subsection_temporal_fusion} discusses our novel approach that integrates information flow between successive frames in the latent space over time, \cf~\Fig~\ref{figure_spatio_temporal_fusion_network}.

\subsection{Pair Network}\label{subsection_pair_network}
\noindent This section introduces our \emph{pair network}, a modified version of~\cite{wang_mvdepthnet_2018} that integrates
additional feature extraction and feature pyramid network (FPN)~\cite{lin_feature_2017} modules into the architecture,~\cf~\Fig\ref{figure_pair_network}. 
The model function $f_{\theta}$ is acquired by assigning $\delta\!=\!1$ in Eq.~\ref{equation_mvs_black_box}, 
\ie, it takes an intrinsic matrix, a reference image at time $t$, one measurement image from time $t\!-\!1$ 
and their poses, and predicts a depth map for the reference frame.
The network consists of five main parts.

\customsubsection{Feature Extraction with FPN.} 
Our feature extraction is based on MnasNet~\cite{tan_mnasnet_2019}, 
which is chosen due to its low-latency.
To increase the receptive field and recover the spatial resolution, it is extended with a feature pyramid  network~(FPN), 
which is shown to be effective for object detection tasks~\cite{lin_feature_2017}. 
MnasNet layers spatially scale down the feature maps until $\scriptstyle{\tfrac{H}{32}\!\times\!\tfrac{W}{32}}$ 
and the FPN recovers the resolution up to $\scriptstyle{\tfrac{H}{2}\!\times\!\tfrac{W}{2}}$. 
As the result of all convolution operations, a feature at the center of the half 
resolution feature map has a receptive field of $304 \!\times\! 304$. 
The output channel size of the FPN is ${C\!H}\!=\!32$. 
The cost volume is constructed using only the feature maps at half resolution, %
while the lower resolution feature maps carried to the encoder with skip connections for additional high-level feature information.

\customsubsection{Cost Volume Construction.} 
We generate a cost volume by employing the traditional plane-sweep stereo~\cite{collins_space-sweep_1996, gallup_real-time_2007} 
with $M\!=\!64$ plane hypotheses, each parameterized by its depth $d_{m}$
and
uniformly sampled in inverse depth space (uniform in pixel space)
within the interval corresponding to 
$d_{near}\!=\!0.25, d_{far}\!=\!20$ meters. 
The cost induced by the $m^{th}$ depth plane is calculated from the pixel-wise correlation 
between the reference feature map $\+F$ and the warped measurement feature map $\+{\tilde{F}}^{m}$. 
Our cost volume $\+V$ is then composed as %
\begin{equation}
V_{i, j, m} = -\innerprod{\+F_{i, j, :}}{\+{\tilde{F}}^{m}_{i, j, :}}/C\!H, \text{\quad for\ }  1\leq m \leq M.
\label{equation_cost_volume_calculation}
\end{equation}

\customsubsection{Cost Volume Encoder-Decoder.} 
The key purpose of the encoder-decoder is to spatially regularize the raw cost volume with a U-Net~\cite{ronneberger_u-net_2015} style architecture. 
The encoder extracts the high-level, global scene information and aggregates the pixel-wise matching costs with the help of the feature maps coming from the feature extraction step. 
The decoder gradually upsamples the high-level encoding to the finer resolutions, 
using the low-resolution inverse depth maps and the skip connections coming from the encoder as guidance. 

\customsubsection{Depth Regression and Refinement.} 
After acquiring the encoding $\+Y$ at the output of each decoder block, we apply one $\text{3}\!\times\!\text{3}$ 
convolution filter $\+w$ and a sigmoid activation $\sigma$.
We pass this tensor to the next decoder block or the refinement block as a guidance, and also regress a depth map from it.
For a pixel location $(i,j)$ in $\widehat{\+D}$, the regression is \vspace{-0.5ex}
\begin{equation}\label{equation_depth_regression}
\widehat{D}_{i,j}\!=\!\left(\!\left(\frac{1}{d_{near}}-\frac{1}{d_{far}}\right)\sigma\left((\+w * \+Y)_{i,j}\right)+\frac{1}{d_{far}}\right)^{-1}\mkern-18mu. \vspace{-0.5ex}
\end{equation}
The finest resolution that the decoder produces is $\scriptstyle{\tfrac{H}{2} \!\times\! \tfrac{W}{2}}$. 
To acquire $\+Y$ at full resolution, we first upsample the output encoding 
of the final decoder block together with the corresponding inverse depth map prediction and concatenate these with the input image. Then, we pass this tensor to two more convolutional layers.

\customsubsection{Loss Function.} 
For our loss function, we accumulate the average $L1$ error over the inverse depth maps at each output resolution considering only valid groundtruth values.

\subsection{Spatio-temporal Fusion}\label{subsection_temporal_fusion}
\begin{figure*}[t]
    \centering
    {\includegraphics[width=0.85\linewidth]{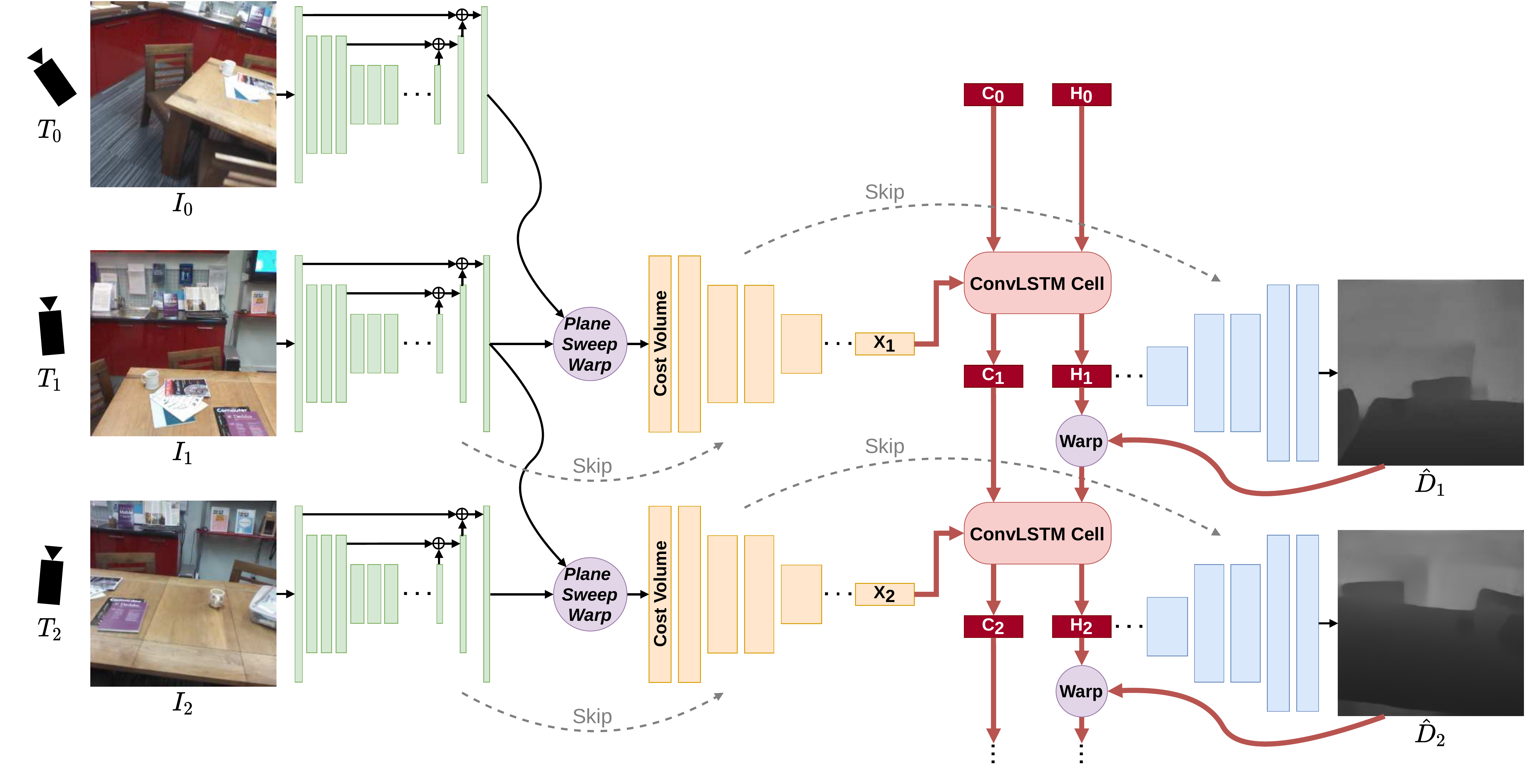}}
    \caption{
        In our fusion approach, the pair network is extended with a ConvLSTM cell placed between the encoder and the decoder. 
        The current frame and the previous frame(s) %
        are used for computing a cost volume. 
        The encoder takes the cost volume and produces a latent encoding at the bottleneck. 
        The current latent encoding then participates in the convolutions in the ConvLSTM cell together with the 
        \emph{warped} hidden state coming from the previous time step forming a Markov chain. 
        After the fusion, the new hidden state is passed through the decoder which outputs the depth predictions in the same way as the pair network.
        }
    \label{figure_spatio_temporal_fusion_network}
\end{figure*}

\noindent We now present our framework that extends our pair network to incorporate knowledge 
about the past into the current prediction when processing video streams. 
In essence, we include a ConvLSTM cell to our network between the encoder and the decoder at the bottleneck to model the spatio-temporal relations, \cf~\Fig~\ref{figure_spatio_temporal_fusion_network}.

Our ConvLSTM cell logic is based on~\cite{palazzi_ndrplzconvlstm_pytorch_2020}, a variant of the original version~\cite{shi_convolutional_2015}.
Let $\+H$ and $\+C$ denote the hidden state and the cell state and $\+X$ denote 
the output of the encoder at the bottleneck, then, the logic is written as
\begingroup
\allowdisplaybreaks
\begin{align}
\+i_{t} &= \sigma\left(\+w_{x i} * \+X_{t} + \+w_{h i} * \+H_{t-1}\right) \nonumber \\
\+f_{t} &= \sigma\left(\+w_{x f} * \+X_{t} + \+w_{h f} * \+H_{t-1}\right) \nonumber \\
\+o_{t} &= \sigma\left(\+w_{x o} * \+X_{t} + \+w_{h o} * \+H_{t-1}\right) \nonumber \\
\+g_{t} &= \text{ELU}\left(\text{layernorm}(\+w_{x g} * \+X_{t} + \+w_{h g} * \+H_{t-1})\right) \nonumber \\
\+C_{t} &= \text{layernorm}(\+f_{t} \odot \+C_{t-1} + \+i_{t} \odot  \+g_{t}) \nonumber \\
\+H_{t} &= \+o_{t} \odot \text{ELU}\left(\+C_{t} \right),
\label{equation_conv_lstm_cell} 
\end{align}
\endgroup
where $*$ denotes convolution, $\odot$ the Hadamard product, $\sigma$ the sigmoid activation %
and $\+w$ are learned convolution filter weights. %
We empirically found that $\text{ELU}$ activation~\cite{clevert_fast_2016} leads to better results than $\tanh$, and Layer Normalization~\cite{ba_layer_2016}, without learnable parameters, %
stabilizes the model by preventing values from growing uncontrollably and enforcing zero mean / unit variance per channel in~$\+C$. %
We also observed that ConvGRU~\cite{DBLP:journals/corr/BallasYPC15} cell, while being a resource-wise cheaper option, perform worse than ConvLSTM when used in our fusion scheme. Hence we opt for the latter. %
For analyses on the recurrent cell choice and the activation-normalization options, please refer to the supplementary material.

\customsubsection{Naive Fusion.}
Ensuring that $\+H$'s and $\+X$'s dimensions are equal (so that the decoder of the pair network can be used without any modification), and letting $\+S$ denote the skip connections from the encoder to the decoder, a simple model for the fusion could be written as
\begingroup
\allowdisplaybreaks
\begin{align}
\+X_{t}, \+S_{t} &= \text{encoding}(\+I_t,\ \+I_{t-1},\ \+T_t,\ \+T_{t-1},\ \+K)\nonumber \\
\+H_{t}, \+C_{t} &= \text{cell}(\+X_t,\ \+{H}_{t-1},\ \+C_{t-1})\nonumber \\
\+{\widehat{D}}_t &= \text{decoding}(\+{H}_{t},\ \+S_{t}).
\label{equation_naive_lstm_fusion}
\end{align}
\endgroup
Note that (\cf \Eq~\ref{equation_conv_lstm_cell}), 
$\+X_{t}$, the output of the encoder at the bottleneck and 
the hidden state $\+{H}_{t-1}$ are directly interacting with each other, 
while being encoded from different viewpoints. 
Under the assumption that $\+X$ and $\+H$ are purpose-wise similar latent representations, 
encapsulating both the visual (image features) and the geometric (cost encoding) information, 
\Eq~\ref{equation_naive_lstm_fusion} forces the ConvLSTM cell to capture the pose-induced image motion between the two encodings,
which can be challenging due to large disparities in near objects, occlusions, rapid rotations, \etc. 

\customsubsection{Proposed Fusion.}
Hence, we find it beneficial to partially account for the viewpoint changes 
while propagating the hidden state $\+H_{t-1}$ to the next time step.
Having access to the past and current camera poses and using the current reconstruction as a proxy, 
we warp the hidden representation 
$\+{H}_{t-1}$ to the current viewpoint to acquire $\+{\widetilde{H}}_{t-1}$. 
Such warping can be implemented as either forward or inverse mapping. 
The former is quite involved~\cite{Niklaus_2020_CVPR} and would require non-trivial 
visibility handling and differentiable rendering, with an additional impact on processing time.
Using bilinear grid sampling~\cite{jaderberg_spatial_2016}, the latter is a fully differentiable and lightweight operation, and lets us keep the overall runtime low.  
To compute the sampling locations,
we estimate a (partial) depth map $\+{\widetilde{D}}_t$ for the current time step \emph{before} starting the forward pass. 
To that end, we project (handling occlusions) a small point cloud, which is estimated from the previous depth prediction $\+{\widehat{D}}_{t-1}$, onto the current camera plane. Then, we sample the hidden representation $\+{H}_{t-1}$ to get $\+{\widetilde{H}}_{t-1}$. 
Here, recall that the input resolution get spatially downscaled by $1/32$, 
while the channel size increases significantly at the bottleneck. 
This relaxes the need for perfect warpings as it enables 
the receptive fields in the ConvLSTM cell to perceive a considerable 
amount of information.
Therefore, we transform the model into
\begingroup
\allowdisplaybreaks
\begin{align}
\widetilde{\+D}_{t} &= \text{projection}(\+{\widehat{D}}_{t-1},\ \+T_{t-1},\ \+T_{t},\ \+K)\nonumber \\
\widetilde{\+H}_{t-1} &= \text{warping}(\+{H}_{t-1},\ \+{\widetilde{D}}_{t})\nonumber \\
\+X_{t}, \+S_{t} &= \text{encoding}(\+I_t,\ \+I_{t-1},\ \+T_t,\ \+T_{t-1},\ \+K)\nonumber \\
\+H_{t}, \+C_{t}  &= \text{cell}(\+X_t,\ \+{\widetilde{H}}_{t-1},\ \+C_{t-1})\nonumber \\
\+{\widehat{D}}_t &= \text{decoding}(\+{H}_{t},\ \+S_{t}).
\label{equation_warp_lstm_fusion}
\end{align}
\endgroup

We argue that our formulation results in an easier learning problem for the ConvLSTM cell, since~\Eq~\ref{equation_warp_lstm_fusion} alleviates the need to
capture the flow of the visual representations along with learning to fuse the encodings. 
We interpret this scheme as the ConvLSTM cell aggregating the prior geometric 
cost and the current geometric cost for the overlapping regions. %

To set the training in motion and stabilize the behaviour, we use the groundtruth depth map $\+D_t$
in place of $\+{\widetilde{D}}_t$ in \Eq~\ref{equation_warp_lstm_fusion}
and switch to the testing strategy only at a late stage, where we solely finetune the cell.
This guides the ConvLSTM first with accurate warpings, and then allows it to adapt to the testing configuration slowly.

\section{Experiments}
\begin{table*}
\centering
\scriptsize
\makeatletter
\newcommand*{\myoverline}[1]{$\overline{\hbox{#1}}\m@th$}
\makeatother
\renewcommand*{\arraystretch}{1.1}
\newcommand{\pullupspace}{-14.5pt}
\newcommand{\roundprecision}{4}
\newcommand{\columnsize}{1.09cm}
\sisetup{round-mode=places, detect-all=true, detect-weight=true, detect-inline-weight=math, group-digits = false, table-number-alignment = center}
    \begin{tabular}{r *{6}{S[table-format = 1.\roundprecision, round-precision=\roundprecision]} | *{4}{S[table-format = 1.\roundprecision, round-precision=\roundprecision]}}
    \toprule
    \phantom{.} & 
    \parbox[c]{\columnsize}{\centering MVDep} & 
    \parbox[c]{\columnsize}{\centering MVDep\\(FT)} &
    \parbox[c]{\columnsize}{\centering DPSNet} & 
    \parbox[c]{\columnsize}{\centering DPSNet\\(FT)} &
    \parbox[c]{\columnsize}{\centering DELTAS} &
    \parbox[c]{\columnsize}{\centering Ours\\(Pair)} &
    \parbox[c]{\columnsize}{\centering NRGBD} &
    \parbox[c]{\columnsize}{\centering GPMVS} & 
    \parbox[c]{\columnsize}{\centering GPMVS\\(FT)} & 
    \parbox[c]{\columnsize}{\centering Ours\\(Fusion)} \\ 
    \cmidrule { 2 - 11 } \vspace{\pullupspace} \\
    \textbf{SCANNET} & & & & & & & & & & \\
    abs             & 0.1953 & 0.1671 & 0.2185 & 0.1607 & 0.1492 & \myoverline{0.1454} & 0.2361 & 0.2027 & 0.1491 & \bfseries 0.1187 \\
    abs-rel         & 0.0970 & 0.0869 & 0.1192 & 0.0828 & 0.0783 & \myoverline{0.0736} & 0.1220 & 0.1088 & 0.0762 & \bfseries 0.0599 \\
    abs-inv         & 0.0621 & 0.0540 & 0.0710 & 0.0530 & 0.0506 & \myoverline{0.0468} & 0.0745 & 0.0669 & 0.0490 & \bfseries 0.0381 \\
    $\delta < 1.25$ & 0.8947 & 0.9252 & 0.8682 & 0.9254 & 0.9383 & \myoverline{0.9459} & 0.8501 & 0.8893 & 0.9396 & \bfseries 0.9654 \\[1pt]
    \cmidrule { 2 - 11 } \vspace{\pullupspace} \\
    \textbf{7-SCENES} & & & & & & & & & & \\
    abs             & 0.2029 & 0.2012 & 0.2486 & 0.1966 & 0.1911 & 0.186  & 0.2143 & 0.1962 & \myoverline{0.1737} & \bfseries 0.1448 \\
    abs-rel         & 0.1157 & 0.1165 & 0.1486 & 0.1148 & 0.1139 & 0.1073 & 0.1312 & 0.1178 & \myoverline{0.1003} & \bfseries 0.0829 \\
    abs-inv         & 0.0732 & 0.0708 & 0.0847 & 0.0729 & 0.0717 & 0.0653 & 0.0756 & 0.0756 & \myoverline{0.0641} & \bfseries 0.0537 \\
    $\delta < 1.25$ & 0.8687 & 0.8768 & 0.8257 & 0.8708 & 0.8821 & 0.8936 & 0.8645 & 0.8723 & \myoverline{0.9034} & \bfseries 0.9380 \\[1pt]
    \cmidrule { 2 - 11 } \vspace{\pullupspace} \\
    \textbf{RGB-D V2} & & & & & & & & & & \\
    abs             & 0.1387 & 0.131  & 0.152  & 0.132  & 0.1581 & 0.1433 & \bfseries 0.1227 & 0.1576 & 0.1275 & \myoverline{0.1256} \\
    abs-rel         & 0.0853 & 0.0846 & 0.1069 & 0.0835 & 0.1098 & 0.0891 & 0.0871 & 0.1029 & \myoverline{0.0773} & \bfseries 0.0765 \\
    abs-inv         & 0.0617 & 0.0637 & 0.0763 & 0.0623 & 0.0811 & 0.0667 & 0.07   & 0.0761 & \bfseries 0.0559 & \myoverline{0.0566} \\
    $\delta < 1.25$ & 0.9417 & 0.9562 & 0.9057 & 0.9422 & 0.9137 & 0.9461 & 0.9373 & 0.9256 & \myoverline{0.9575} & \bfseries 0.9707 \\[1pt]
    \cmidrule { 2 - 11 } \vspace{\pullupspace} \\
    \textbf{TUM RGB-D} & & & & & & & & & & \\
    abs             & 0.2902* & 0.326  & 0.2958* & 0.3045 & 0.3525 & 0.3535 & 0.3185 & \bfseries{0.2443}* & 0.2938 & \myoverline{0.2878} \\
    abs-rel         & 0.1172* & 0.1259 & 0.1335* & 0.1184 & 0.1273 & 0.1247 & 0.1209 & \myoverline{0.1038}* & 0.1103 & \bfseries 0.0975 \\
    abs-inv         & 0.0643* & 0.0715 & 0.0758* & 0.0722 & 0.0753 & 0.0736 & 0.0681 & \myoverline{0.0604}* & 0.0643 & \bfseries 0.0551 \\
    $\delta < 1.25$ & 0.8597* & 0.8304 & 0.8314* & 0.8347 & 0.8177 & 0.8249 & 0.8402 & \bfseries{0.8892}* & 0.8563 & \myoverline{0.8845} \\[1pt]
    \cmidrule { 2 - 11 } \vspace{\pullupspace} \\
    \textbf{ICL-NUIM} & & & & & & & & & & \\
    abs             & \bfseries 0.1392 & 0.1574 & 0.1695 & \myoverline{0.1491} & 0.1953 & 0.1771 & 0.173  & 0.1667 & 0.1558 & 0.1496 \\
    abs-rel         & \bfseries 0.0581 & 0.0637 & 0.0756 & 0.0612 & 0.0844 & 0.0723 & 0.0764 & 0.0709 & 0.0623 & \myoverline{0.0587} \\
    abs-inv         & \myoverline{0.0305} & 0.0328 & 0.0367 & 0.0359 & 0.0458 & 0.0402 & 0.0365 & 0.0356 & 0.0323 & \bfseries 0.0297 \\
    $\delta < 1.25$ & 0.9468 & \myoverline{0.9477} & 0.9348 & 0.9401 & 0.9192 & 0.9331 & 0.9318 & 0.9381 & 0.9442 & \bfseries 0.9571 \\[1pt]
    \hline
    \end{tabular}
    \caption{
    Performance on: \textbf{i.} ScanNet test set, \textbf{ii.} 13 sequences from 7-Scenes , \textbf{iii.} 8 sequences from RGB-D Scenes V2, \textbf{iv.} 13 sequences from TUM RGB-D and \textbf{v.} 4 sequences from Augmented ICL-NUIM. 
    Except Neural RGBD that uses 4 measurement frames, all evaluated methods use a \emph{single} measurement frame.
    \emph{FT} denotes finetuned on ScanNet. 
    \emph{Bold} is the best score, \emph{overline} indicates the second best score. 
    The vertical line separates video agnostic (\emph{left}) from video aware (\emph{right}) methods.
    *~the method is already trained on most of the test frames. 
    }
    \label{table_depth_prediction_quantitative}
\end{table*}
\noindent
Our model is implemented in PyTorch and trained using one NVIDIA GTX 1080Ti GPU, Intel i7-9700K CPU. 
The fusion network is trained with a mini-batch size of 4 and a subsequence length of 8.
Input image size is 256$\times$256 for which we crop the original image to a square, then scale. Additional technical details and the exact training procedure are provided in the supplementary.
In summary, we first train the pair network independently and use the  weights to partially initialize our fusion network. 
We start by training the cell and the decoder, which are randomly initialized, and then gradually unfreeze the other modules.
Finally, we finetune only the cell while warping the hidden states with the predictions. 
To prevent overfitting of the regression part,
we employ a variant of geometric scale augmentation~\cite{wang_mvdepthnet_2018}, 
with a random geometric scale factor between $0.666$ and $1.5$. 

\subsection{Dataset}
\label{subsection_dataset}
\noindent Training our spatio-temporal fusion network based on short-term memory demands longer input sequences.
Therefore, we opt for the ScanNet~\cite{dai_scannet_2017} dataset's official training split to train and validate our models.
For testing, we use ScanNet's 100 sequence official test split and a diverse collection of sequences, without large dynamic objects, from other datasets. We select 13 sequences from~\cite{glocker_real-time_2013}, 8 from~\cite{lai_unsupervised_2014}, 13 from~\cite{sturm_benchmark_2012}, and 4 from~\cite{handa_benchmark_2014, sungjoon_choi_robust_2015} (\cf \Tab~\ref{table_depth_prediction_quantitative}). Altogether, there are around 31K images in the test set.
The official distribution of~\cite{glocker_real-time_2013} does not supply aligned color and depth images, 
so we use the rendered depth maps provided by~\cite{brachmann_learning_2018} for evaluation.

\subsection{Frame Selection}
\label{subsection_frame_selection}
\noindent
View selection, \ie picking the right measurement frame for a reference frame, is an often overlooked but crucial aspect for balancing triangulation quality, matching accuracy, and view frustum overlap. 
To that end, we utilize the pose-distance measure proposed by~\cite{hou_multi-view_2019} 
\vspace{-0.5ex}
\begin{equation}
dist\left[\+T_\textrm{rel}\right] = \sqrt{\left\|\mathbf{t}_\textrm{rel}\right\|^{2}+\frac{2}{3} \operatorname{tr}\left(\mathbb{I}-\mathbf{R}_\textrm{rel}\right)},\vspace{-0.5ex}
\label{equation_pose_distance_measure}
\end{equation}
where $\+T_\textrm{rel}\!=\![\ \+R_\textrm{rel}\ |\ \+t_\textrm{rel}\ ]$ %
denotes the relative pose between two cameras. 
We empirically found that a maximum pose-distance of 0.35~$\pm$~0.05 and a translation 
of $10$~cm~$\pm$~$5$~cm ensures sufficient baselines and image overlaps for indoor scenes, while being realistic for potential hand-held operation cases.
For training, we augment our dataset further by sampling 
subsequences with thresholds from the prescribed range, 
where each consecutive frame obeys a given threshold.
For testing, we simulate an online system that buffers the last 30 keyframes and updates the buffer frequently. %
If the pose-distance from the most recent keyframe is above 0.1, we add a new one, and rank the buffered keyframes~\wrt
\vspace{-1ex}
\begin{align}
penalty(\+T_{rel}) &= \alpha(\|\+t_{rel}\| - 0.15)^2 + \frac{2}{3}\operatorname{tr}\left(\mathbb{I}-\mathbf{R}_{rel}\right) \nonumber \\
\alpha & =
\begin{cases}
5.0\ \ \text{if}\ \|\+t_{rel}\| \leq 0.15\\
1.0\ \ \text{if}\ \|\+t_{rel}\| > 0.15,
\end{cases}     
\label{equation_penalty}
\end{align}
which prefers relative camera distances of $15$~cm %
to select the desired number of measurement frames.

\subsection{Comparison with Existing Methods}
\label{subsection_comparison_with_existing}
\noindent
We compare our method with five state-of-the-art, learning-based MVS approaches: 
MVDepthNet~\cite{wang_mvdepthnet_2018}, GP-MVS~\cite{hou_multi-view_2019} (in online mode of operation), DPSNet~\cite{im_dpsnet_2019}, 
Neural RGBD~\cite{liu_neural_2019}, and DELTAS~\cite{sinha_deltas_2020}. 
Recall that MVDepthNet, DPSNet, DELTAS and our pair network %
do not exploit the sequential structure of the video input  %
and only consider a given amount of frames at a time. 
In contrast, GP-MVS, Neural RGBD, and our proposed fusion approach are 
tailored for operating on video streams. %
We finetune MVDepthNet, GP-MVS and DPSNet on the ScanNet training set
for up to 200K iterations and use the model with the best validation loss. 
We evaluate these both before and after finetuning. %

\customsubsection{Quantitative Evaluation.}
\begin{figure*}[tb]
\renewcommand*{\arraystretch}{1.1}
  \begin{center}
    \begin{minipage}{0.725\textwidth}
    \begin{flushleft}
    \captionsetup{justification=raggedright, singlelinecheck=false}
    \newcommand{\roundprecision}{4}
    \scriptsize
    \sisetup{round-mode=places, detect-all=true, detect-weight=true, detect-inline-weight=math, group-digits = false, table-number-alignment = center}
      \begin{tabular}{r|*{2}{*{3}{S[table-format = 1.\roundprecision, round-precision=\roundprecision]}|}*{3}{S[table-format = 1.\roundprecision, round-precision=\roundprecision]}}
        \toprule
        \multirow{2}{*}{\phantom{.}} &
          \multicolumn{3}{c|}{\underline{1 Measurement Frame}} &
          \multicolumn{3}{c|}{\underline{2 Measurement Frames}} &
          \multicolumn{3}{c}{\underline{3 Measurement Frames}} \\
          \rule{0pt}{8pt} \phantom & {abs} & {abs-rel} & {abs-inv} & {abs} & {abs-rel} & {abs-inv} & {abs} & {abs-rel} & {abs-inv} \\
          \midrule
        MVDepthNet & 0.177  & 0.0892 & 0.0548 & 0.1757 & 0.0877 & 0.0542 & 0.1739 & 0.0872 & 0.0544\\
        GPMVS & 0.1589 & 0.0786 & 0.0498 & 0.1588 & 0.0779 & 0.0496 & 0.1609 & 0.08   & 0.0513\\
        DELTAS & 0.1659 & 0.0837 & 0.0533 & 0.16   & 0.0798 & 0.0505 & 0.1562 & 0.0776 & 0.0492\\
        Ours (Pair) & 0.1614 & 0.0783 & 0.0492 & 0.1587 & 0.0757 & 0.0475 & 0.1537 & 0.0729 & 0.0457 \\
        Ours (Fusion) & \bfseries 0.1318 & \bfseries 0.0634 & \bfseries 0.0397 & \bfseries 0.1320 & \bfseries 0.0619 & \bfseries 0.0387 & \bfseries 0.1298 & \bfseries 0.0609 & \bfseries 0.0381 \\
        \bottomrule
      \end{tabular}
      \captionof{table}{Effect of using multiple measurement frames. Our view selection heuristic is used.}\vspace{4ex}
      \label{table_number_of_measurement_frames}
      \begin{tabular}{r|*{2}{*{3}{S[table-format = 1.\roundprecision, round-precision=\roundprecision]}|}*{3}{S[table-format = 1.\roundprecision, round-precision=\roundprecision]}}
        \toprule
        \multirow{2}{*}{\phantom{.}} &
          \multicolumn{3}{c|}{\underline{Our Keyframe Buffer}} &
          \multicolumn{3}{c|}{\underline{Every 10\textsuperscript{th} Frame}} &
          \multicolumn{3}{c}{\underline{Every 20\textsuperscript{th} Frame}} \\
          \rule{0pt}{8pt} \phantom & {abs} & {abs-rel} & {abs-inv} & {abs} & {abs-rel} & {abs-inv} & {abs} & {abs-rel} & {abs-inv} \\
          \midrule
        MVDepthNet    & 0.1757 & 0.0877 & 0.0542 & 0.1841 & 0.094  & 0.0594 & 0.1865 & 0.0989 & 0.0628\\
        GPMVS         & 0.1588 & 0.0779 & 0.0496 & 0.1682 & 0.0838 & 0.0545 & 0.1799 & 0.0942 & 0.061 \\
        DELTAS        & 0.1600   & 0.0798 & 0.0505 & 0.1694 & 0.0858 & 0.0555 & 0.1712 & 0.0885 & 0.0574\\
        Ours (Pair)   & 0.1587 & 0.0757 & 0.0475 & 0.1686 & 0.0814 & 0.0530  & 0.1583 & 0.0770  & 0.0500\\
        Ours (Fusion) & \bfseries 0.1320 & \bfseries 0.0619 & \bfseries 0.0387 & \bfseries 0.1398 & \bfseries 0.0663 & \bfseries 0.0426 & \bfseries 0.1376 & \bfseries 0.0663  & \bfseries 0.0429\\
        \bottomrule
      \end{tabular}
      \captionof{table}{Effect of applying our view selection heuristic. Two measurement frames are used.}
      \label{table_keyframe_selection}
    \end{flushleft}
    \end{minipage}%
    \begin{minipage}{0.275\textwidth}
    \centering
    \includegraphics[width=0.85\linewidth]{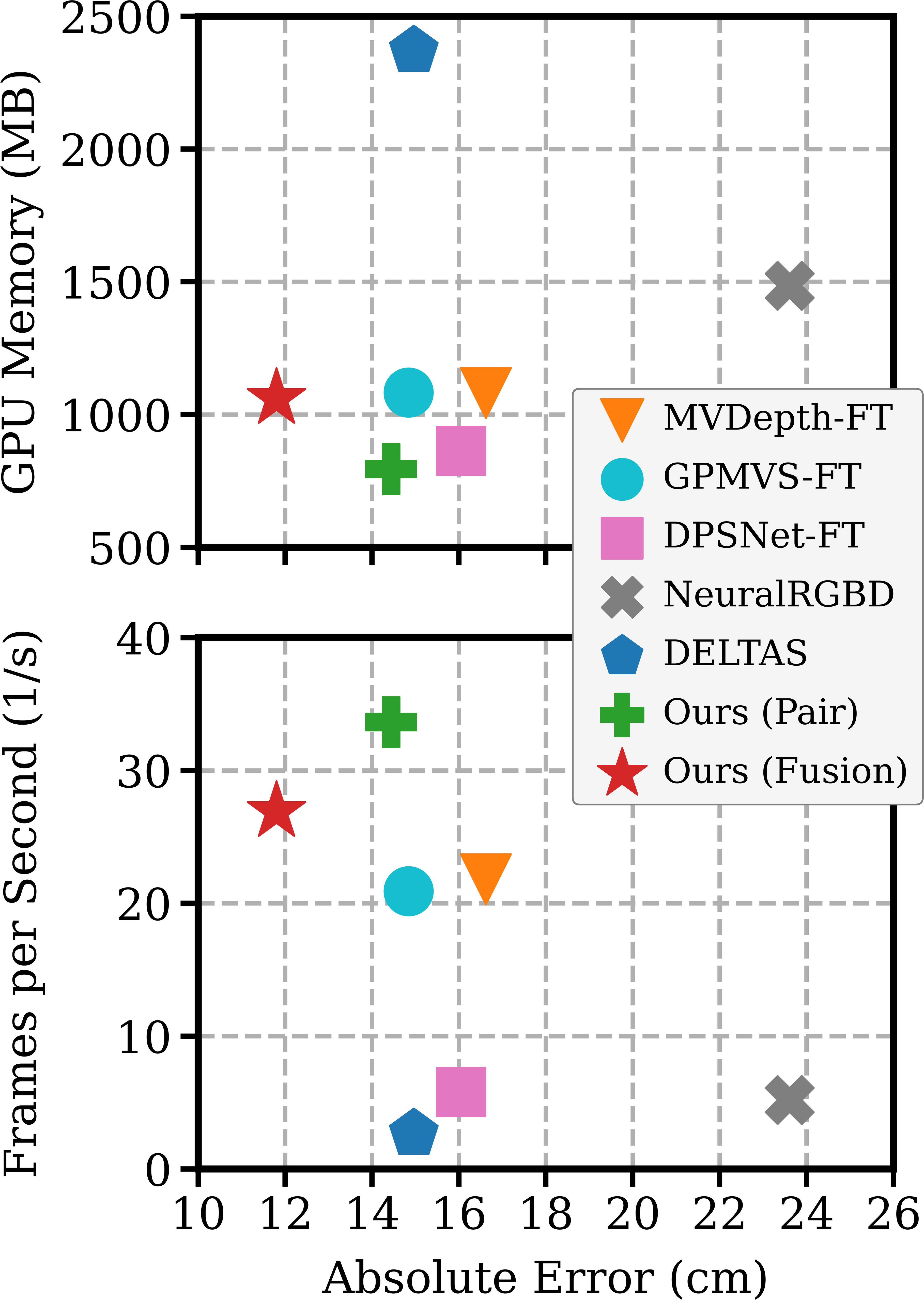}\vspace{2pt}
    \captionof{figure}
    {
        Speed and memory consumption in relation to depth prediction performance on ScanNet.
    }
    \label{figure_inference_memory_vs_accuracy}
    \end{minipage}
  \end{center}\vspace{-3ex}
\end{figure*}
We use the following standard metrics~\cite{eigen_depth_2014, sinha_deltas_2020} to acquire quantitative results:
mean absolute depth error (abs), mean absolute relative depth error (abs-rel), mean absolute inverse depth error (abs-inv), %
and inlier ratio with threshold 1.25 ($\delta \!<\! 1.25$). 
Since most of our competitors are limited to a minimum depth of 0.5 meters, 
we do not consider the groundtruth measurements below this threshold for evaluation.
For a fair comparison on full field of views, we run the inference for our models at $320\!\times\!256$ resolution without cropping.
We acquire the predictions of each method at their native input resolutions by following their suggested scaling, 
then upsample the predictions with nearest neighbour interpolation to the original size ($640\!\times\!480$) before calculating the metrics.

\Tab~\ref{table_depth_prediction_quantitative} summarizes our results.
Our pair network performs on par with state-of-the-art competitors. With spatio-temporal fusion, we outperform the existing methods in $70\%$ of all the metrics. 
Notably, our spatio-temporal fusion approach steadily improves the performance of the backbone across all test sets. 
For instance, when averaged over all test sets, we get $19.3\%$ improvement in absolute inverse depth error, \cf \Tab~\ref{table_number_of_measurement_frames}.
In comparison, the finetuned GP-MVS improves over its MVDepthNet backbone by only $9.1\%$.  %
Note that, both fusion strategies can extend many other existing approaches to leverage the past information. 
Being trained on ScanNet's official training split, DELTAS performs well on the ScanNet test sequences, 
but their performance drops below many of the other methods on the rest of the test sets. 
In contrast, our fusion approach generalizes well and delivers across the board. Even on TUM RGB-D, it is competitive
against methods that are already trained on most of the \emph{test} frames. %
Overall, our fusion model outperforms the best competitor, finetuned GP-MVS by a large margin, 
\eg by $20.3\%$ in absolute inverse depth error (\cf \Tab~\ref{table_number_of_measurement_frames}). 

\begin{figure*}[t]
\centering
{\includegraphics[clip, trim={0 0 0 0.15cm},
width=\linewidth]{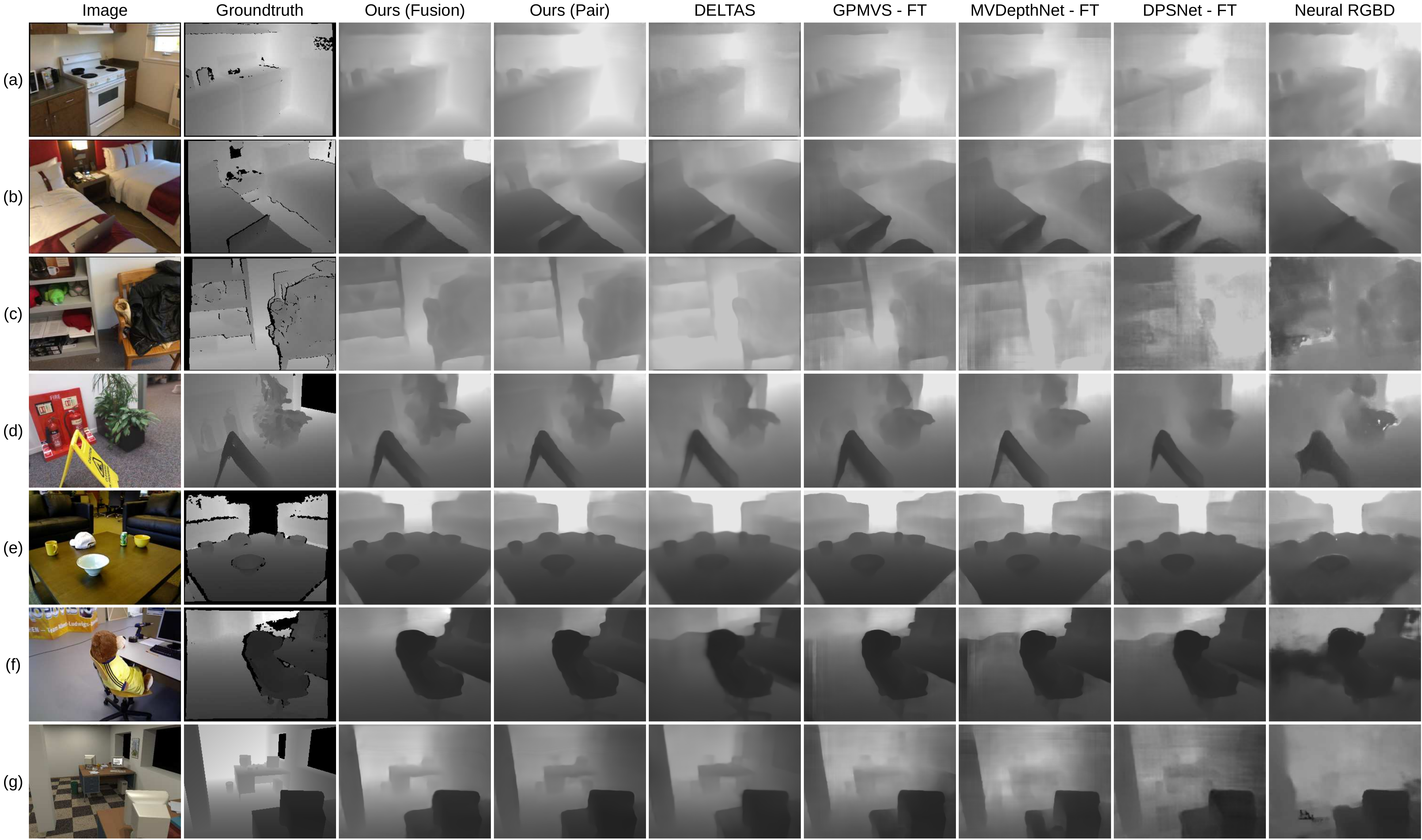}}
\caption{Example depth predictions from all test sets. Examples (a), (b) and (c) are taken from ScanNet, (d) from 7-Scenes, (e) from RGB-D Scenes V2, (f) from TUM RGB-D SLAM, and (g) from Augmented ICL-NUIM.}
\label{figure_qualitative_depth_maps_1}
\end{figure*}
\begin{figure*}[tb]
\centering
{\includegraphics[width=\linewidth]{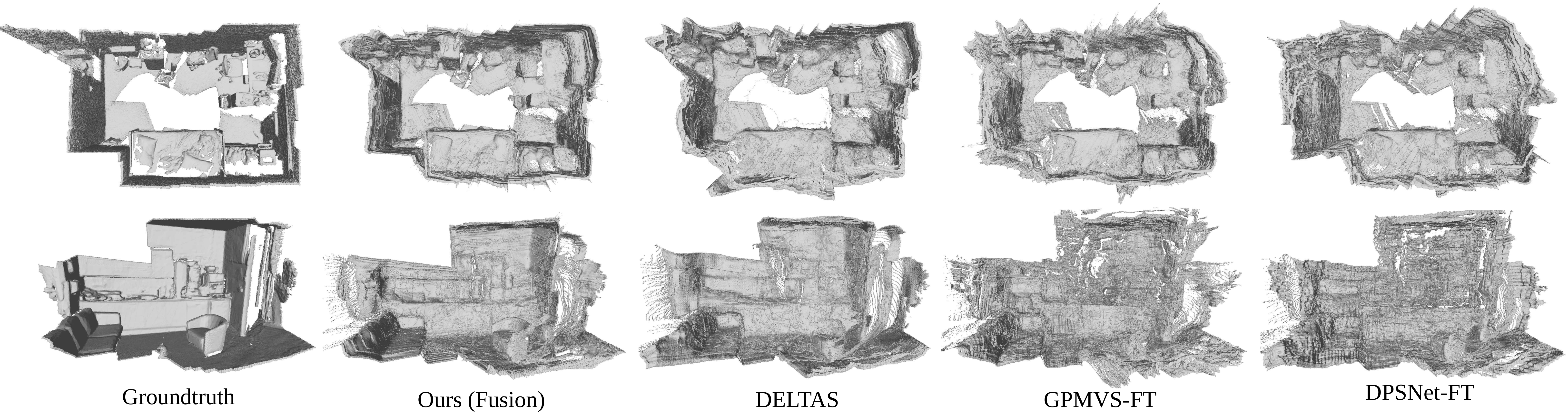}}
\caption{TSDF reconstructions from ScanNet (top) and 7-Scenes (bottom). Our fusion approach produces less noisy and geometrically more consistent depth predictions that result in accurate 3D reconstructions, \eg straight walls, perpendicular corners, armchair, sofa.}
\label{figure_tsdf_reconstructions_1}
\end{figure*}
\customsubsection{Qualitative Evaluation.}
\Fig~\ref{figure_qualitative_depth_maps_1} compares the methods qualitatively. 
Reconstructions of Neural RGBD appear noisy and blurry. 
MVDepthNet and DPSNet suffer from prevalent gridding artifacts, visible in~(b),~(c)~and~(g).
Being based on MVDepthNet, GP-MVS cannot completely eliminate the artifact,~\eg visible in~(c). 
The depth maps of DELTAS are visually pleasant but blurry around edges at depth discontinuities. 
Overall, our fusion approach appears to be the one that is closest to the groundtruth and visibly improves upon our pair network. 
For instance, the corner of the kitchen in~(a), the shelves in~(c), arm of the sofa in~(e), or the far wall in~(g) are assigned more coherent depth values. 

\Fig~\ref{figure_tsdf_reconstructions_1} shows TSDF reconstructions, acquired using the toolbox from~\cite{zeng_3dmatch_2017}, from ScanNet and 7-Scenes.
The increased consistency among the depth predictions of our fusion network result in less noisy reconstructions, the walls appear flat and geometry of the scene is preserved the best.
For instance, only our method allows to identify the armchair in the bottom row.

\customsubsection{Runtime and Memory.}
\Fig\ref{figure_inference_memory_vs_accuracy} relates the absolute error to the memory consumption and average runtime of a single forward pass, measured on NVIDIA GTX 1080Ti. Exact numbers are provided in the supplementary.
Our methods are the fastest and most accurate, especially our fusion approach delivers unmatched prediction quality at a high frame rate. %

\begin{table}[t]
\scriptsize
    \renewcommand*{\arraystretch}{1.1}
    \begin{center}
        \begin{widetable}{\columnwidth}{ r | c | c | c }
            \toprule
            \phantom
            &
            abs
            & 
            abs-rel
            & 
            abs-inv
            \\
            \midrule
            Pair Network & 0.1614 & 0.0783 & 0.0492\\
            Fusion without warping (\Eq~\ref{equation_naive_lstm_fusion}) & 0.1495 & 0.0697 & 0.0436\\
            Fusion with warping (\Eq~\ref{equation_warp_lstm_fusion}) & \textbf{0.1318} & \textbf{0.0634} & \textbf{0.0397}  \\
            \bottomrule
        \end{widetable}
        \caption{Effect of the proposed hidden state propagation scheme.}
        \label{table_effect_of_warping}
    \end{center}\vspace{-3ex}
\end{table}

\subsection{Ablation Studies}
\label{subsection_ablation_studies}
\noindent
For our ablation studies, we average the performance over all test sequences. 
We present only the most important results here and more studies can be found in the supplementary.

\customsubsection{Propagating the Hidden State by Warping.} 
\Tab~\ref{table_effect_of_warping} shows that 
already the naive fusion scheme achieves on average about 10\% improvement over the pair network, while the proposed propagation scheme delivers a similar gain on top.

\customsubsection{Number of Measurement Frames.}
MVS methods typically use several measurement frames to gain robustness and coverage at the expense of time and memory. Our fusion approach is orthogonal to this idea, and one can similarly construct and average multiple cost volumes.
As \Tab~\ref{table_number_of_measurement_frames} shows, additional views tend to improve the performance, but it can stagnate after a number of measurement frames. While DELTAS and our models steadily benefit from more measurements, the abs-inv error of MVDepthNet and GPMVS, for instance, increases at three views.

\customsubsection{Frame Selection.}
\Tab~\ref{table_keyframe_selection} shows that even a simple view selection heuristic enables consistently better predictions, compared to the common sampling rates (about 10\% on average). 
Note that, our frame selection heuristic affects all methods in a similar manner. 
Still, all evaluated MVS methods, including ours, are robust enough
to also work under naive sampling of every 10\textsuperscript{th} or 20\textsuperscript{th} frame in a video.

\section{Conclusion}
\noindent

\noindent In this work, we tackle the problem of predicting depth maps from posed-video streams. 
Our approach exploits the temporally structured input and can be operated 
as an online multi-view stereo system, while estimating very accurate depth maps in real-time. %
Starting from a lightweight stereo backbone, we integrate a memory cell that acts as a fusion module %
and agglomerates the information obtained within the bottleneck encodings of our backbone over time. 
To remedy the viewpoint dependence of the fused encodings, we explicitly transform the 
hidden state of the ConvLSTM cell, while propagating it through time. 
Our proposed approach outperforms the existing state-of-the-art methods in 
most of the evaluation metrics and generalizes well to all considered test sets. 
We also achieve a significantly faster inference time than 
all competitors, 
while keeping a low memory consumption. 

\begin{adjustwidth}{-0.75ex}{-0.75ex}

{\small
\bibliographystyle{bib/ieee_fullname}
\bibliography{bib/abbreviation_short,bib/bibliography}
}

\end{adjustwidth}
\clearpage
\appendix
\twocolumn[%
\begin{center}
      {\Large \bf Supplementary Material for\\DeepVideoMVS: Multi-View Stereo on Video\\with Recurrent Spatio-Temporal Fusion \par} \vspace{3ex}
\end{center}]

\section{Training Details}
\noindent We use an image size of 256\!$\times$\!256 with cropping and scaling. Since the original image resolution is 640\!$\times$\!480 in the ScanNet dataset, we crop the image from the left and right equally to acquire an image size of 480\!$\times$\!480, then downscale to 256\!$\times$\!256.
During training, we compute the cost volume for a reference frame using one measurement frame. 

\customsubsection{Pair Network Training.}
We use the Adam optimizer with $\beta_1\!=\!0.9$ and $\beta_2\!=\!0.999$, and a constant learning rate of $1\mathrm{e}{-4}$. The pair network is trained with a mini-batch size of 14 for 600K iterations in total. We load the pretrained weights for the MnasNet layers, which are supplied by PyTorch, and freeze these layers for the first 200K iterations. We predict depth maps for both of the input images at each forward pass, \ie, after the shared feature extraction modules, we consider a given feature map once as the reference feature map and once as the measurement feature map.

\customsubsection{Fusion Network Training.}
The fusion network is trained with a subsequence length of 8 and a mini-batch size of 4.  As the initial weights for the feature extraction layers, the feature pyramid network (FPN) and the encoder, we use the checkpoint saved at 100K iterations of the pair network training. We freeze the parameters for these modules and train only the ConvLSTM cell and the decoder for 25K iterations, which are randomly initialized. Next, we add the FPN and the encoder weights among the trainable parameters and train for another 25K iterations. Then, we unfreeze the MnasNet layers and train the whole network up to 1000K iterations, while validating and saving checkpoints frequently for early stopping. During these iterations, we use the groundtruth depth map to warp the hidden state. Finally, we load the best checkpoint and finetune only the cell for another 25K iterations with a learning rate of $5\mathrm{e}{-5}$ while warping the hidden states with the predicted depth maps. We do not allow the gradients to flow through the depth map prediction that warps the hidden state. As our attempts suggest, allowing this gradient flow introduces a complex causal relationship among the predictions and the gradients explode when the training experiences a large loss in an arbitrary subsequence.

\customsubsection{Data Augmentation.}
We use several data augmentation techniques. These are applying slight changes to the color space of the images, randomly reversing the order of the frames in a subsequence, and randomly changing the geometric scale of the subsequences. Also, we apply random horizontal flips to the cost volume and the extracted features during pair network training to increase the diversity of the cases that the encoder and the decoder experience. For the geometric scale augmentation, we choose the default sampling interval of the scaling factor as $[0.666, 1.5]$. When the training pipeline requests a new subsequence from the data pipeline, we adjust these bounds according to the minimum and the maximum depth measurements in the subsequence. This ensures that even after applying a random geometric scaling, the cost volume can still contain a correct local extremum for every pixel. The sampled scale factor is multiplied with the groundtruth depth values and the translation columns of the camera poses in an entire subsequence.

\section{Evaluation Metrics}
\noindent For a predicted depth map, the metrics are calculated as:
\begin{enumerate}[label=\bfseries\roman*.]
    \item abs = $\dfrac{1}{N}\sum\limits_{i=1}^{N}|d_{i}-\hat{d}_{i}|$
    \item abs-rel = $\dfrac{1}{N} \sum\limits_{i=1}^{N}\dfrac{|d_{i}-\hat{d}_{i}|}{d_{i}}$
    \item abs-inv = $\dfrac{1}{N} \sum\limits_{i=1}^{N}\left|\dfrac{1}{d_{i}}-\dfrac{1}{\hat{d}_i}\right|$
    \item inlier ratio = $\dfrac{1}{N} \sum\limits_{i=1}^{N}\mathds{1}\left[\dfrac{d_i}{1.25} < \hat{d}_i < 1.25\!\times\!d_i\right]$
\end{enumerate}
\noindent where $\hat{d}_i$ and $d_i$ denote the predicted depth value and the groundtruth depth value for a given pixel $i$ that has a valid depth measurement, $N$ is the number of pixels that have valid depth measurements, and $\mathds{1}$ is the indicator function.

\section{Inference Time and Memory Consumption}
\noindent The average duration of a single forward pass and the GPU memory consumption for the methods in comparison are given in Tab.~\ref{table_inference_time_memory}. The measurements are taken on a workstation with NVIDIA GTX 1080Ti GPU. We do not consider the time spent by the CPU or the transfer time of the data to/from the GPU. We fix the input image sizes for all methods to $320\!\times\!256$, start recording the times after the first 100 forward passes (warm start), and always start a method when the CPU and GPU temperature are below 40$^{\circ}$C. We run all the methods with a mini-batch size of 1, \ie, the simulation of predicting a single depth map for the current time step. For DELTAS, we replace the PyTorch's singular value decomposition with a custom function that the authors suggest to increase the inference speed. However, note that we could not reproduce the inference times that the authors of DELTAS report as around 90 milliseconds on NVIDIA Titan RTX GPU.
\begin{table}[h]
\scriptsize
    \renewcommand*{\arraystretch}{1.3}
    \begin{center}
        \begin{tabular}{r | c | c | c}
            \toprule
            \phantom
            & 
            \multicolumn{1}{p{1.55cm}|}{\centering\ Time(ms)\ $\boldsymbol{\downarrow}$}
            & 
            \multicolumn{1}{p{1.15cm}|}{\centering\ \ FPS\ $\boldsymbol{\uparrow}$}
            &
            \multicolumn{1}{p{2.05cm}}{\centering\ Memory(MB)\ $\boldsymbol{\downarrow}$}
            \\
            \midrule
            \arrayrulecolor{table_gray}
            \multirow{1}{*}{MVDepth\hspace{0pt}}              & 45.84 & $\sim\!21.8$ & 1081\\
                                                              \hline
            \multirow{1}{*}{GPMVS\hspace{0pt}}                & 47.81 & $\sim\!20.9$ & 1083\\
                                                              \hline
            \multirow{1}{*}{DPSNet\hspace{0pt}}               & 172.62 & $\sim\!5.8$ & 863\\
                                                              \hline
            \multirow{1}{*}{NRGBD\hspace{0pt}}                & 192.43 & $\sim\!5.2$ & 1485\\
                                                              \hline
            \multirow{1}{*}{*DELTAS\hspace{0pt}}              & 374.78 & $\sim\!2.7$ & 2371\\
            \arrayrulecolor{black}
            \hline
            \multirow{1}{*}{Ours (Pair)}                       & 29.72 & $\sim\!33.7$ & 795\\
                                                              \arrayrulecolor{table_gray}
                                                              \hline
            \multirow{1}{*}{Ours (Fusion)}                     & 37.14 & $\sim\!26.9$ & 1061\\
            \arrayrulecolor{black}
            \bottomrule
        \end{tabular}
        \caption{Mean inference time and GPU memory consumption of all methods. Inference time is averaged over 300 forward passes. All methods are run with an image size of $320\!\times\!256$ and use one measurement frame, except Neural RGBD that requires minimum two measurement frames. *~the authors report faster times.\vspace{-3ex}}
        \label{table_inference_time_memory}
    \end{center}
\end{table}

\section{Evaluation Set Details}
\noindent We use four real and one synthetic dataset to evaluate the methods. All of the datasets provide RGB-D videos at $640\!\times\!480$ depth image resolution and 3D camera poses. The selected sequences are as follows:
\begin{itemize}
    \item \textbf{ScanNet}\cite{dai_scannet_2017}: All 100 official test sequences, \ie, scene0707 to scene0806.
    \item \textbf{7-Scenes}\cite{glocker_real-time_2013}: chess-01, chess-02, fire-01, fire-02, head-02, office-01, office-03, pumpkin-03, pumpkin-06, redkitchen-01, redkitchen-02, stairs-02, stairs-06.
    \item \textbf{RGB-D Scenes V2}\cite{lai_unsupervised_2014}: scene-01, scene-02, scene-05, scene-06, scene-09, scene-10, scene-13, and scene-14.
    \item \textbf{TUM RGB-D SLAM}\cite{sturm_benchmark_2012}: fr1-desk, fr1-plant, fr1-room, fr1-teddy, fr2-desk, fr2-dishes, fr2-large-no-loop, fr3-cabinet, fr3-long-office-household, fr3-nostructure-notexture-far, fr3-nostructure-texture-far, fr3-structure-notexture-far, fr3-structure-texture-far.
    \item \textbf{Augmented ICL-NUIM}\cite{handa_benchmark_2014, sungjoon_choi_robust_2015}: livingroom1, livingroom2, office1, office2.
\end{itemize}

\begin{figure*}[t]
    \centering
    \includegraphics[width=0.75\linewidth]{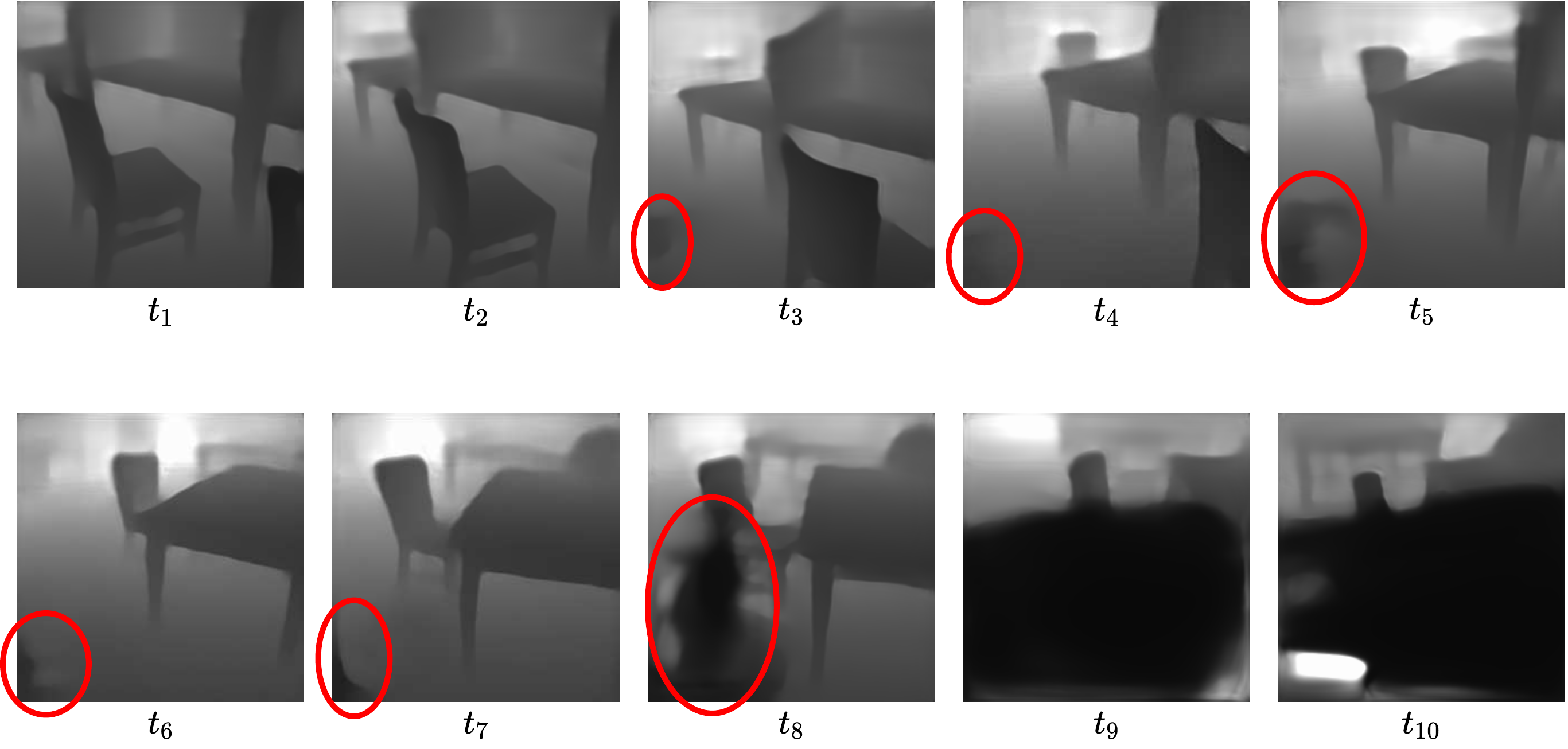}
    \caption{Placing ELU activations without any normalization in the ConvLSTM cell causes stability issues. The noise starting at the lower left corner of the hidden state (thus the prediction), gets propagated through time and grows with the subsequent convolution operations.}
    \label{figure_explosion_of_elu}
\end{figure*}

\section{Additional Ablation Studies}
\noindent In this section, we provide additional ablation studies. We present the results acquired on our \emph{validation split} of the ScanNet dataset. These experiments were conducted at different stages of this work. Each subsection is self-contained, \ie the models in a given subsection are trained with identical training and data pipelines. However, the results cannot be meaningfully compared in between subsections, since they may employ different training strategies.

\begin{figure}[H]
\centering
{\includegraphics[width=0.78\linewidth]{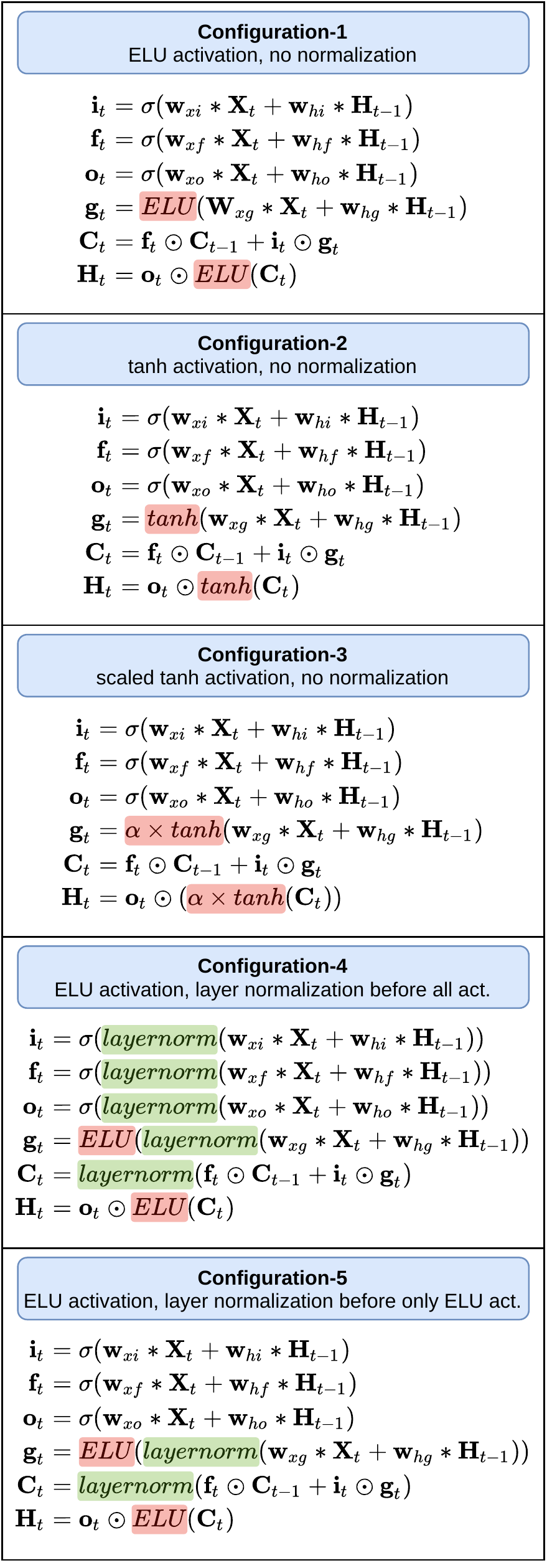}}
\caption{Various activation and normalization options for the ConvLSTM cell. Configuration-5 delivers the best depth prediction performance, while maintaining a stable behaviour.}
\label{figure_activation_function_table}
\end{figure}

\customsubsection{Activation and Normalization in the ConvLSTM Cell.} There are many activation and normalization options that can be included in a ConvLSTM cell. The options that showed significant effects are provided in Fig.~\ref{figure_activation_function_table} and the corresponding depth prediction performances are given in Tab.~\ref{table_different_activation_normalization}.

\begin{table}[t]
\scriptsize
    \renewcommand*{\arraystretch}{1.3}
    \begin{center}
        \begin{tabular}{ l | c | c | c | c | c}
            \toprule
            \phantom
            &
            abs
            & 
            abs-rel
            & 
            abs-inv
            & 
            noise test
            &
            loop test
            \\
            \midrule
            Configuration-1\hspace{1pt} & 0.1267 & 0.0580 & 0.0344 & \xmark & \xmark\\
            Configuration-2\hspace{1pt} & 0.1321 & 0.0601 & 0.0351 & \cmark & \cmark\\
            Configuration-3\hspace{1pt} & 0.1272 & 0.0590 & 0.0351 & \cmark & \cmark\\
            Configuration-4\hspace{1pt} & 0.1236 & 0.0578 & 0.0347 & \cmark & \cmark\\
            Configuration-5\hspace{1pt} & \textbf{0.1206} & \textbf{0.0567} & \textbf{0.0339} & \cmark & \cmark\\
            \bottomrule
        \end{tabular}
        \caption{\centering Evaluation of different activation and normalization options inside the ConvLSTM cell. \vspace{-2.5ex}}
        \label{table_different_activation_normalization}
    \end{center}
\end{table}

Placing ELU activations enables the range of values to be similar at the output of the encoder and the output of the ConvLSTM cell. Whereas, tanh limits the range of the hidden state to $[-1, 1]$. During training, we observe that ELU (Configuration-1) achieves a better learning curve and provides better validation scores than tanh (Configuration-2). However, since we validate on short subsequences, similar to the training subsequences, we can not observe the side effects of the ELU activation during training. When the validation is run on long sequences, we observe that naively placing ELU activations causes two issues. Both of them are due to the output range $[-1, \infty]$ of ELU activation, which is asymmetric and unbounded on the positive side.

The first issue is that, if the model fails to produce a ``good" hidden state at an arbitrary time, and instead computes a slightly noisy state, the noise gets propagated
and grows incrementally, causing whole predictions to diverge over time. This issue is depicted in Fig. \ref{figure_explosion_of_elu}. It can occur at any arbitrary time in a sequence, both with and without the warping of the hidden state. Randomly introducing positive additive noise to the hidden state consistently reproduced this issue and served as a test of robustness for all the activation and normalization configurations. We call this the ``noise test". The second issue is observed while simulating online multi-view stereo captures with tens of thousands of frames, done by running the inference on the sequences in our validation split that include pose loops. We call this the ``loop test". With ELU activation and no normalization, the cell state ($\+C_{t}$) is mostly increasing. In very long sequences, the values in the states ($\+H_{t}$, $\+C_{t}$) eventually reach very large values, which destabilizes the model and the predictions diverge similar to the first issue.

Scaling the tanh activation function (Configuration-3) with a learnable $\alpha$, which the training ends up assigning 4.459, adjusts the numerical range of the hidden state and results in better depth predictions than unscaled tanh. However, it can not achieve the accuracy of the unstable Configuration-1. Whereas, Configuration-4 gives similar results to Configuration-1, while also passing the stability test and the loop test. In this configuration, the sparse nature of ELU activations are kept while the layer normalization ensures that the cell state always have a zero mean and unit variance per channel, and the values do not grow uncontrollably. We further improve the performance by removing the layer normalizations before the sigmoid activations as in Configuration-5. Placing layer normalizations before the sigmoid activations dictate that, in $\+i$, $\+f$, $\+o$ gate tensors, roughly half of the values are always below 0.5 and half of them are above 0.5. This unnecessarily constrains the learning, thus removing them results in better depth predictions.

\customsubsection{ConvLSTM vs. ConvGRU.}
We compare ConvGRU and ConvLSTM cells as the recurrent cell choice for our proposed fusion module. ConvLSTM cell logic is given in Eq.~\ref{equation_conv_lstm_cell} in the paper. By applying the principles learned from the ablation studies on activation and normalization options, ConvGRU cell is set up as
\begin{align}
\+u_{t} &= \sigma (\+w_{xu} * \+X_{t} + \+w_{hu} * \+H_{t-1}) \nonumber \\
\+r_{t} &= \sigma (\+w_{xr} * \+X_{t} + \+w_{hr} * \+H_{t-1}) \nonumber \\
\+o_{t} &= \text{ELU} (\text{layernorm}(\+w_{xo} * \+X_{t} + \+w_{ho} * (\+H_{t-1} \odot \+r_{t}))) \nonumber \\
\+H_{t} &= \text{layernorm}(\+u_{t} \odot \+H_{t-1} + (1 - \+u_{t}) \odot \+o_{t}).
\label{equation_conv_gru_cell}
\end{align}
We test the performances of these cells while warping the hidden states during propagation. Tab.~\ref{table_comparison_convlstm_convgru} presents the study results where ConvLSTM outperforms its counterpart by 5.6\% in  abs-inv error. We speculate that warping the \emph{sole} hidden state in ConvGRU hinders the memory functionality. Since each warping operation removes information from non-overlapping view frustums of consecutive frames externally, such info gets lost irrecoverably in the ConvGRU case. Whereas, with ConvLSTM, we take advantage of the two internal states and manipulate only the hidden state. 

\begin{table}[h]
\scriptsize
    \renewcommand*{\arraystretch}{1.3}
    \begin{center}
        \begin{tabular}{ l | c | c | c }
            \toprule
            \phantom
            &
            abs
            & 
            abs-rel
            & 
            abs-inv
            \\
            \midrule
            Fusion with ConvLSTM (Eq.~\ref{equation_conv_lstm_cell}) and Warping & \textbf{0.1192} & \textbf{0.0565} & \textbf{0.0340}\\
            Fusion with ConvGRU (Eq.~\ref{equation_conv_gru_cell}) and Warping & 0.1248 & 0.0594 & 0.0359\\
            \bottomrule
        \end{tabular}
        \caption{\centering Comparison of ConvLSTM and ConvGRU cell performances when placed in the proposed fusion module.\vspace{-2ex}}
        \label{table_comparison_convlstm_convgru}
    \end{center}
\end{table}

\customsubsection{Finetuning the Cell.}
In Tab.~\ref{table_effect_of_finetuning}, B already provides a high performance, which shows that the model does not require pixel-perfect warpings while propagating the hidden states. Nevertheless, finetuning the ConvLSTM cell with the test time strategy reduces the discrepancy between using the groundtruth depth and predicted depth, \cf relative differences between A, B and C, D. Surprisingly, we also observe a slight improvement from A to C.
\begin{table}[h]
\scriptsize
    \renewcommand*{\arraystretch}{1.3}
    \begin{center}
        \begin{tabular}{l | c | c | c | c }
            \toprule
            \phantom
            &
            Warp With
            &
            abs
            & 
            abs-rel
            & 
            abs-inv
            \\
            \midrule
            Before Finetuning (\textbf{A}) & Groundtruth & 0.1192 & 0.0565 & 0.0340\\
            Before Finetuning (\textbf{B}) & Prediction  & 0.1207 & 0.0573 & 0.0346\\
            After Finetuning  (\textbf{C}) & Groundtruth & 0.1189 & 0.0563 & 0.0337\\
            After Finetuning  (\textbf{D}) & Prediction  & 0.1191 & 0.0564 & 0.0338
            \\
            \bottomrule
        \end{tabular}
        \caption{\centering Effect of finetuning the cell while warping the hidden states with predictions instead of groundtruths.\vspace{-2ex}}
        \label{table_effect_of_finetuning}
    \end{center}
\end{table}

\customsubsection{Skip Connections from Feature Pyramid to Encoder.}
As presented, we pass multi-scale image features to the cost volume encoder by placing several skip connections from the feature pyramid network. In order to measure the effect of these connections, we remove the low resolution skip connections and keep only the half resolution feature map coming from the feature pyramid, \ie, single-skip design. As shown in Tab.~\ref{table_effect_skipping_not_skipping}, despite achieving similar results in a pair network, multi-skip design is around 4.5\% better in
the abs-inv metric when placed in a fusion network. We conjecture that we mostly benefit from a secondary effect rather than the primary motivation of these connections which is to guide the encoder with strong cues about the image content at each level. The secondary effect is the established phenomenon that skip connections ease the training and smooth the loss landscape\cite{li_visualizing_2018}. Thus, the multi-skip design help the model reach a better loss minimum in the presence of a ConvLSTM (which typically complicates the loss landscape) by providing more ``gradient highways".
\begin{table}[h]
\scriptsize
    \renewcommand*{\arraystretch}{1.3}
    \begin{center}
        \begin{tabular}{ l | c | c | c }
            \toprule
            \phantom
            &
            abs
            & 
            abs-rel
            & 
            abs-inv
            \\
            \midrule
            Pair Network - Multi-Skip Design \hspace{1pt}       & 0.1441 & 0.0695 & 0.0427\\
            Pair Network - Single-Skip Design \hspace{1pt}      & 0.1444 & 0.0692 & 0.0426\\
            Fusion Network - Multi-Skip Design \hspace{1pt}     & 0.1192 & 0.0565 & 0.0340\\
            Fusion Network - Single-Skip Design \hspace{1pt}    & 0.1245 & 0.0589 & 0.0355\\
            \bottomrule
        \end{tabular}
        \caption{\centering Effect of placing multiple skip connections from the feature pyramid to the cost volume encoder.\vspace{-2ex}}
        \label{table_effect_skipping_not_skipping}
    \end{center}
\end{table}

\customsubsection{Accuracy vs. Time Trade-off of the Fusion.} We show the effectiveness of our proposed fusion in an alternative way in Tab.~\ref{table_comparison_with_more_depth_planes}, where we assume a fixed budget for the inference time. We re-train the pair network with the number of sweeping planes $M$=88 to increase the sweep resolution while compensating the additional runtime of the fusion. The increase in the depth resolution improves the pair network's performance as expected, but we only see 1.2\% improvement in abs-inv on our ScanNet validation split. The proposed fusion, with a similar overhead of computational time, results in 20.4\% improvement. This demonstrates the high skew in the accuracy vs. time trade-off.
\begin{table}[ht]
\scriptsize
    \renewcommand*{\arraystretch}{1.3}
    \begin{center}
        \begin{tabular}{ l | c | c | c | c}
            \toprule
            \phantom
            &
            abs
            & 
            abs-rel
            & 
            abs-inv
            &
            time (ms)
            \\
            \midrule
            Pair Network with $M$=64    & 0.1441 & 0.0695 & 0.0427 & 29.72 \\
            Pair Network with $M$=88    & 0.1429 & 0.0688 & 0.0422 & 37.56 \\
            Fusion Network with $M$=64  & 0.1192 & 0.0565 & 0.0340 & 37.14 \\
            \bottomrule
        \end{tabular}
        \caption{\centering Comparison between increasing the plane sweep resolution and employing the proposed fusion mechanism.\vspace{-2ex}}
        \label{table_comparison_with_more_depth_planes}
    \end{center}
\end{table}

\section{Additional Qualitative Results and \\Supplementary Video}
\begin{figure*}
    \centering
    \includegraphics[width=0.9\linewidth]{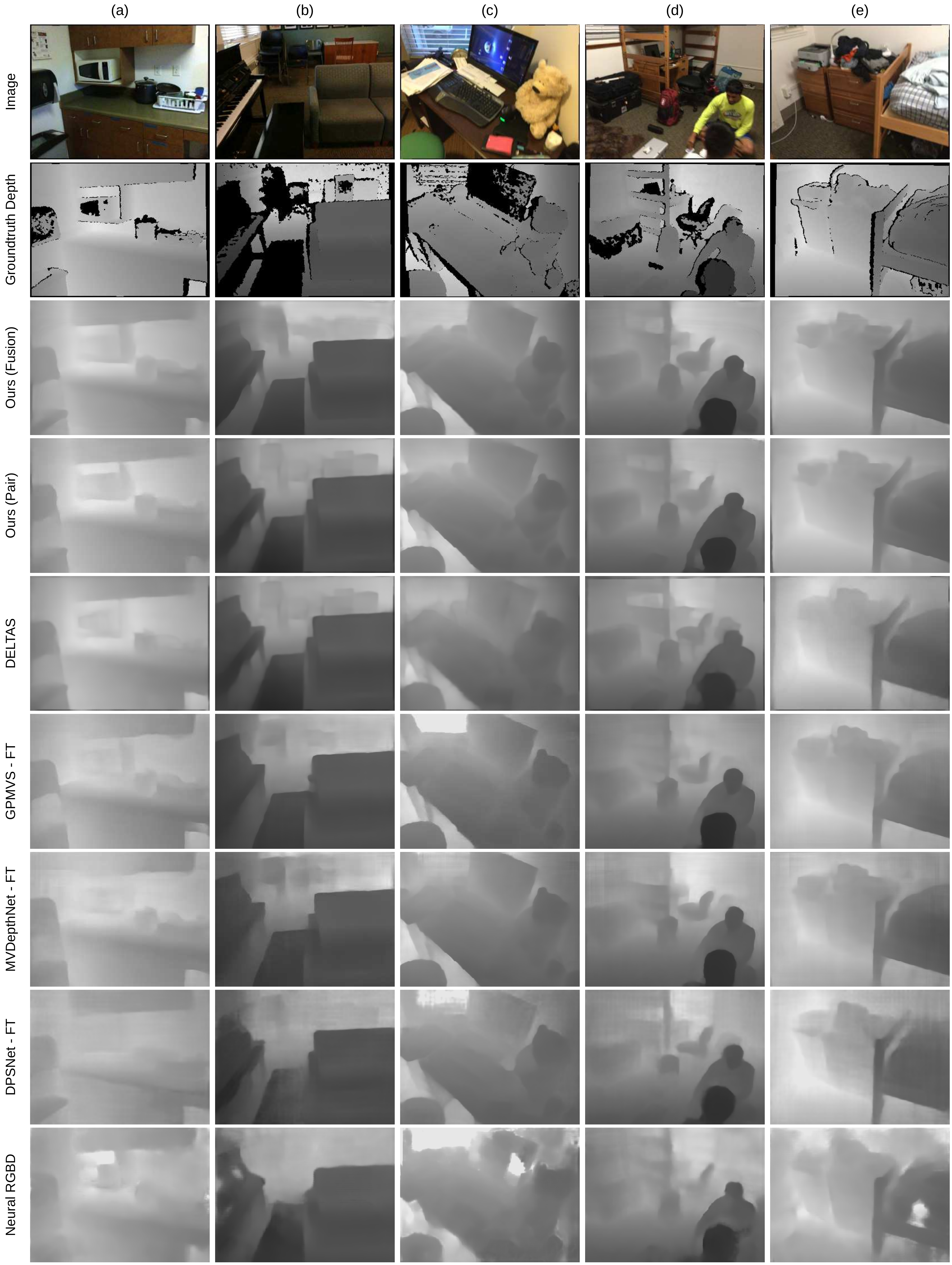}
    \caption{\centering Example depth predictions from ScanNet.}
    \label{figure_supplementary_qualitative_1}
\end{figure*}

\begin{figure*}
    \centering
    \includegraphics[width=0.9\linewidth]{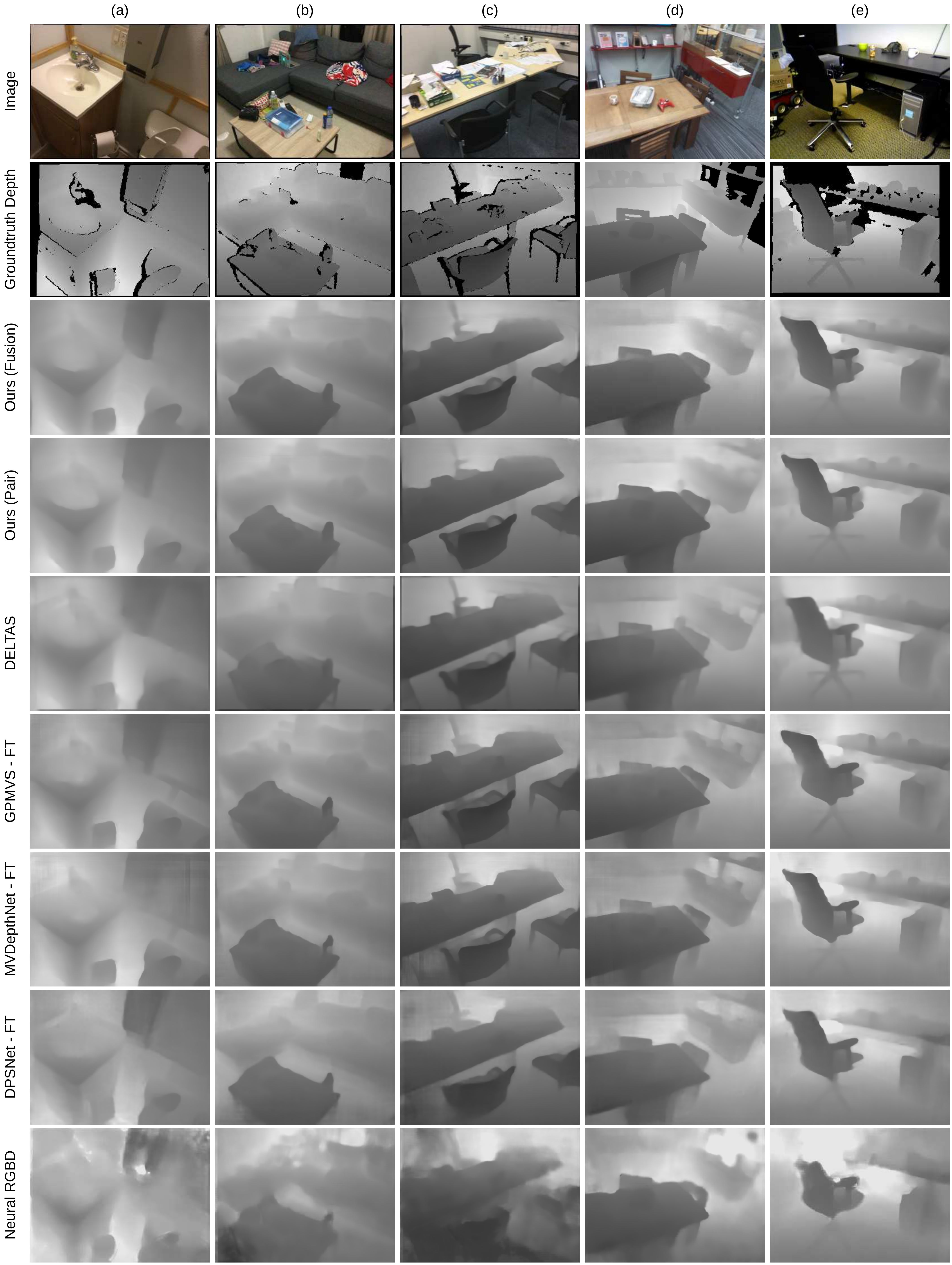}
    \caption{\centering Example depth predictions from ScanNet, 7-Scenes and RGB-D Scenes V2.}
    \label{figure_supplementary_qualitative_2}
\end{figure*}

\noindent In \Fig\ref{figure_supplementary_qualitative_1} and \Fig\ref{figure_supplementary_qualitative_2}, we provide additional example depth predictions. The same trends we discussed in the main paper are observed in these examples too. Our pair network can already output plausible depth predictions, and the proposed fusion module improves the coherency of the depth values for an image leveraging the past information. This effect is better observed in the supplementary video, \small\url{https://ardaduz.github.io/deep-video-mvs/miscellaneous/deep-video-mvs-supplementary-video.mp4}\normalsize, in which we demonstrate the predicted depth map sequences and the resulting TSDF reconstructions of several indoor scenes. Our spatio-temporal fusion network produces predictions with less flickering effects in between time steps, it achieves a superior geometric consistency than our pair network and the existing methods. We noticeably output more consistent depth predictions for the planar surfaces throughout a sequence which gets reflected as smooth reconstructions of such surfaces.

\vspace*{\fill}
\vspace*{\fill}

\vspace*{\fill}
\vspace*{\fill}
\clearpage

\end{document}